\documentclass[letterpaper,journal]{IEEEtran}
\usepackage{amsmath,amsfonts}
\usepackage{algorithmic}
\usepackage{algorithm}
\usepackage{array}
\usepackage[caption=false,font=normalsize,labelfont=sf,textfont=sf]{subfig}
\usepackage{textcomp}
\usepackage{stfloats}
\usepackage{url}
\usepackage{verbatim}
\usepackage{graphicx}
\usepackage{multirow}
\usepackage{colortbl}
\usepackage{xcolor}
\usepackage{booktabs}
\usepackage{cite}
\begin{document}

\title{Selective Noise Suppression and Discriminative Mutual Interaction for Robust Audio-Visual Segmentation}

\author{Kai Peng$^{\#}$, Yunzhe Shen$^{\#}$, Miao Zhang$^{*}$, ~\IEEEmembership{Member, IEEE}, Leiye Liu, Yidong Han, Wei Ji, Jingjing Li, \\Yongri Piao$^{*}$, \IEEEmembership{Member, IEEE} and Huchuan Lu, \IEEEmembership{Fellow, IEEE}
\thanks{This paper was produced by the IEEE Publication Technology Group. They are in Piscataway, NJ. Manuscript received April 19, 2021; revised August 16, 2021. This work was supported in part by the National Natural Science Foundation of China under Grant 62372080, Grant 62376050, and Grant U22B2052, and in part by the Natural Science Foundation of Liaoning Province under Grant 2024-MSBA-24. (\# is equal contribution, * is corresponding author)}
\thanks{Kai Peng, Yunzhe Shen, Miao Zhang, Leiye Liu, Yidong Han, Yongri Piao, and Huchuan Lu are with Dalian University of Technology, Dalian 116024, China (e-mail: happypk@mail.dlut.edu.cn; 1079006\allowbreak460@mail.dlut.edu.cn; miaozhang@dlut.edu.cn; leiyeliu@mail.dlut.edu.cn; 13730\allowbreak982770@mail.dlut.edu.cn; yrpiao@dlut.edu.cn; lhchuan@dlut.edu.cn).}
\thanks{Wei Ji is with the School of Medicine, Yale University, New Haven,
CT 06520 USA (e-mail: wei.ji@yale.edu).}
\thanks{Jingjing Li is with the Department of Electrical and Computer
Engineering, University of Alberta, Edmonton, AB T5V 1A4, Canada(e-mail: jingjin1@ualberta.ca).}
}

\markboth{Journal of \LaTeX\ Class Files,~Vol.~14, No.~8, August~2021}%
{Shell \MakeLowercase{\textit{et al.}}: A Sample Article Using IEEEtran.cls for IEEE Journals}


\maketitle

\begin{abstract}
The ability to capture and segment sounding objects in dynamic visual scenes is crucial for the development of Audio-Visual Segmentation (AVS) tasks. While significant progress has been made in this area, the interaction between audio and visual modalities still requires further exploration. In this work, we aim to answer the following questions: How can a model effectively suppress audio noise while enhancing relevant audio information? How can we achieve discriminative interaction between the audio and visual modalities? To this end, we propose SDAVS, equipped with the Selective Noise-Resilient Processor (SNRP) module and the Discriminative Audio-Visual Mutual Fusion (DAMF) strategy. The proposed SNRP mitigates audio noise interference by selectively emphasizing relevant auditory cues, while DAMF ensures more consistent audio-visual representations. Experimental results demonstrate that our proposed method achieves state-of-the-art performance on benchmark AVS datasets, especially in multi-source and complex scenes. \textit{The code and model are available at \textcolor{blue}{\url{https://github.com/happylife-pk/SDAVS}}.}
\end{abstract}


\section{Introduction}
\IEEEPARstart{T}{he} Audio-Visual Segmentation (AVS) task aims to achieve precise localization and segmentation of sounding objects within visual scenes~\cite{zhou2022audio}, holding substantial potential for autonomous driving~\cite{10337780} and video surveillance~\cite{4066991}. 

Recent research has enhanced AVS performance through two main approaches: audio-prompt learning~\cite{gao2024avsegformer,liu2024annotation} and cross-modal fusion~\cite{huang2023discovering,li2023catr}. The former utilizes audio signals as prompts to fine-tune visual models, while the latter integrates multi-modal features at various stages to localize sounding objects.

\begin{figure*}[!t]
    \centering
    \includegraphics[width=\textwidth]{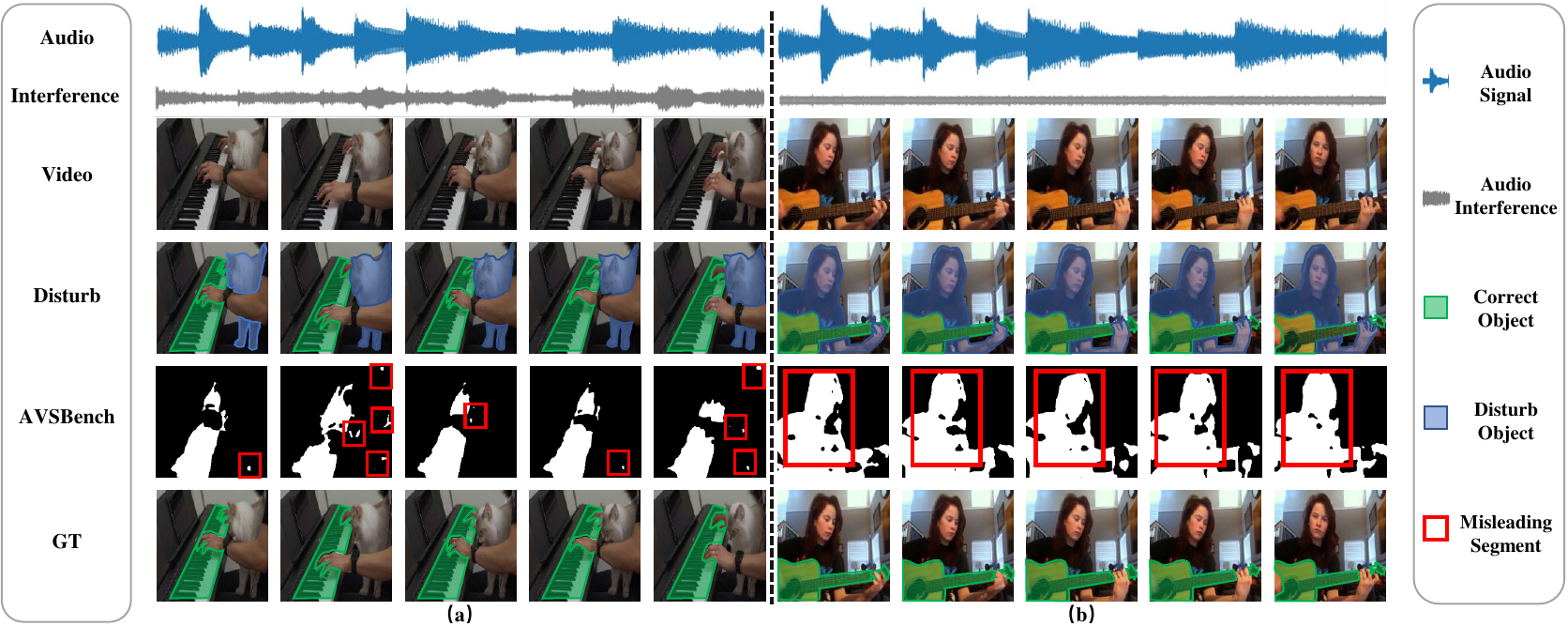}
    \caption{In the first scene (a), the piano is the actual sounding object. However, during the piano performance, the interaction between the cat and the person interferes with the audio signal. In the second scene (b), the guitar is the true sounding object, while the person remains silent. Existing methods struggle to distinguish the correct sounding object.}
    \label{Motivation}
\end{figure*}

Despite their significant success, AVS tasks face two major issues. First, beyond silent visual interference, audio signals contain substantial intrinsic background noise. In this case, using audio features as prompt can cause noise propagation, spreading interference globally throughout the network. Meanwhile, visual information may struggle to adequately emphasize the crucial audio cues, misleading inaccurate sounding object segmentation. For example, silent objects are incorrectly identified as sounding objects, as shown in Fig.~\ref{Motivation}(a). Second, there exists a notable disparity between the audio and visual information. As the visual information provides abundant appearance and spatial cues but does not identify sound sources, while the audio information contains the sound-producing object but struggles with spatial localization. Due to the semantic gap, visual features may highlight silent objects while audio signals struggle to pinpoint precise locations, causing the two modalities to generate conflicting attention maps that interfere with one another, as shown in Fig.~\ref{Motivation}(b).

To address these challenges, we focus on two core objectives: mitigating audio noise interference while extracting essential cues, and enabling discriminative mutual interaction between modalities. To this end, we propose SDAVS, comprising the Selective Noise-Resilient Processor (SNRP) and the Discriminative Audio-Visual Mutual Fusion (DAMF) module.

Specifically, we introduce SNRP to address the noise interference issue. Distinguished from prior methods that primarily focus on visual feature alignment to avoid silent object interference, SNRP targets the substantial intrinsic background noise embedded within audio signals. Designed before the modal fusion stage, it compresses audio information across spatial and channel dimensions to extract high-density semantic representations. By progressively leveraging correlations across modalities, SNRP effectively filters out irrelevant disturbances to prevent noise propagation, ensuring that correlated auditory cues are preserved for subsequent integration.

To effectively address the inherent perception inconsistency issue between the audio and visual modalities, we propose the DAMF module. Initially, DAMF captures rich and efficient representations across spatial-temporal-channel domains. Crucially, DAMF fosters perceptual consensus by establishing a bidirectional mutual guidance mechanism. This allows each modality to discriminatively refine the other by adaptively emphasizing correlated semantic components. By leveraging complementary strengths—rich spatial cues from vision and semantic reliability from audio, DAMF ensures that the consensus is firmly anchored to the correct sounding object, mitigating potential contradictions. Our contributions can be summarized into three aspects:
\begin{itemize}
    \item We introduce the SNRP module, which suppresses irrelevant audio noise while emphasizing correlated auditory cues for more accurate audio–visual integration.
    \item We propose the DAMF strategy, which enhances cross-modal consistency by progressively matching the perception regions of the audio and visual modalities.
    \item We conduct comprehensive quantitative and visualization experiments on AVSBench (S4, MS3, AVSS) and VPO datasets, where SDAVS achieves state-of-the-art performance. In particular, it improves $\mathcal{J}\&\mathcal{F}_m$ on MS3 by +2.6 points, demonstrating its robustness and generalization in complex multi-source and noisy scenarios.
\end{itemize}

\section{Related Work}
Audio-visual (AV) learning has progressed significantly in recent years~\cite{morgado2020learning,nagrani2021attention,hou2022audio}, transitioning from unsupervised correspondence learning to a wide range of downstream tasks that leverage both audio and visual modalities. Among them, Audio-Visual Localization (AVL) and Audio-Visual Segmentation (AVS) have garnered substantial research attention in recent years, primarily stemming from their practical importance in real-world scenarios.
\subsection{Audio-Visual Localization}
Audio-Visual Localization (AVL) aims to localize the visual regions of a video that correspond to the sound source(s). It plays a central role in various applications such as human-computer interaction, surveillance, and assistive technologies.

Early approaches~\cite{arandjelovic2017look,owens2016visually} adopt a weakly supervised paradigm due to the scarcity of annotated datasets available for this task, learning spatial correspondences between audio and visual modalities using classification-based or contrastive objectives. Later works such as AVOL-Net and SoundNet~\cite{senocak2018learning} introduce spatial attention maps to highlight regions in the image corresponding to the sounding object. However, these models generally struggle in complex scenes involving multiple sound sources, occlusion, or background noise. Recent studies~\cite{tian2018audio,9615027,9712233,10856377,10109178} explore transformer-based architectures that model long-range cross-modal interactions, enabling finer localization under more challenging settings. Moreover, temporal modeling has been introduced to capture the dynamic nature of sound sources, improving performance in continuous video streams.

Despite these advances, AVL still largely focuses on region-level localization, lacking precise spatial resolution at the pixel level. This limitation motivates the development of tasks like AVS, which require detailed mask prediction rather than coarse heat-maps or bounding boxes.
\subsection{Audio-Visual Segmentation}
Audio-Visual Segmentation (AVS) extends the scope of AVL by requiring pixel-wise segmentation masks of sound-producing objects in the visual domain, which presents additional challenges in terms of spatial precision, temporal coherence, and multi-source disambiguation. 

Existing supervised audio-visual segmentation (AVS) methods can generally be grouped into two main categories: methods based on fusion~\cite{gao2024avsegformer,liu2024annotation,ma2024stepping,wang2024ref,liu2024bavs} and methods based
on audio-prompt ~\cite{huang2023discovering,li2023catr,ling2023hear,liu2023audio,seon2024extending}. Among the fusion-based works,~\cite{zhou2022audio} introduced a multi-stage integration mechanism that combines audio cues with multi-scale visual features to guide segmentation. Extending this direction, CATR ~\cite{li2023catr} proposed a unified architecture that captures both temporal dependencies and cross-modal fusion, while~\cite{hao2024improving} explored a bidirectional generation strategy, further enhancing segmentation quality by better exploiting inter-modal interactions. Furthermore, AVSAC~\cite{chen2024bootstrapping} addressed modality imbalance caused by unidirectional and insufficient integration of audio cues. It introduced a bidirectional audio-visual decoder integrated bridges to enhance audio representations. Moreover, C3N~\cite{10812843} globally align audio-visual semantics through attention mechanisms, mitigating inconsistencies in noisy environments within fusion-based paradigms. Likewise, temporal and multi-modal Mamba models are explored to delineate sounding objects in video streams, emphasizing robust fusion in complex scenarios~\cite{10891410}. On the other hand, prompt-based methods such as AVSegFormer~\cite{gao2024avsegformer}, CQFormer~\cite{10979212} and GAVS~\cite{wang2024prompting} use audio queries as prompts to decode the fused audio-visual features. GAVS, in particular, exploits rich visual context to build a model capable of generalizing to zero-shot and few-shot conditions. 

Despite the remarkable success of these methods, the noise interference and less discriminative fusion between modalities affect the quality of the segmentation, which is crucial for improving the interpretability and robustness of AVS models, especially in complex scenes.
\begin{figure*}[!t]
  \centering
  \includegraphics[width=\textwidth]{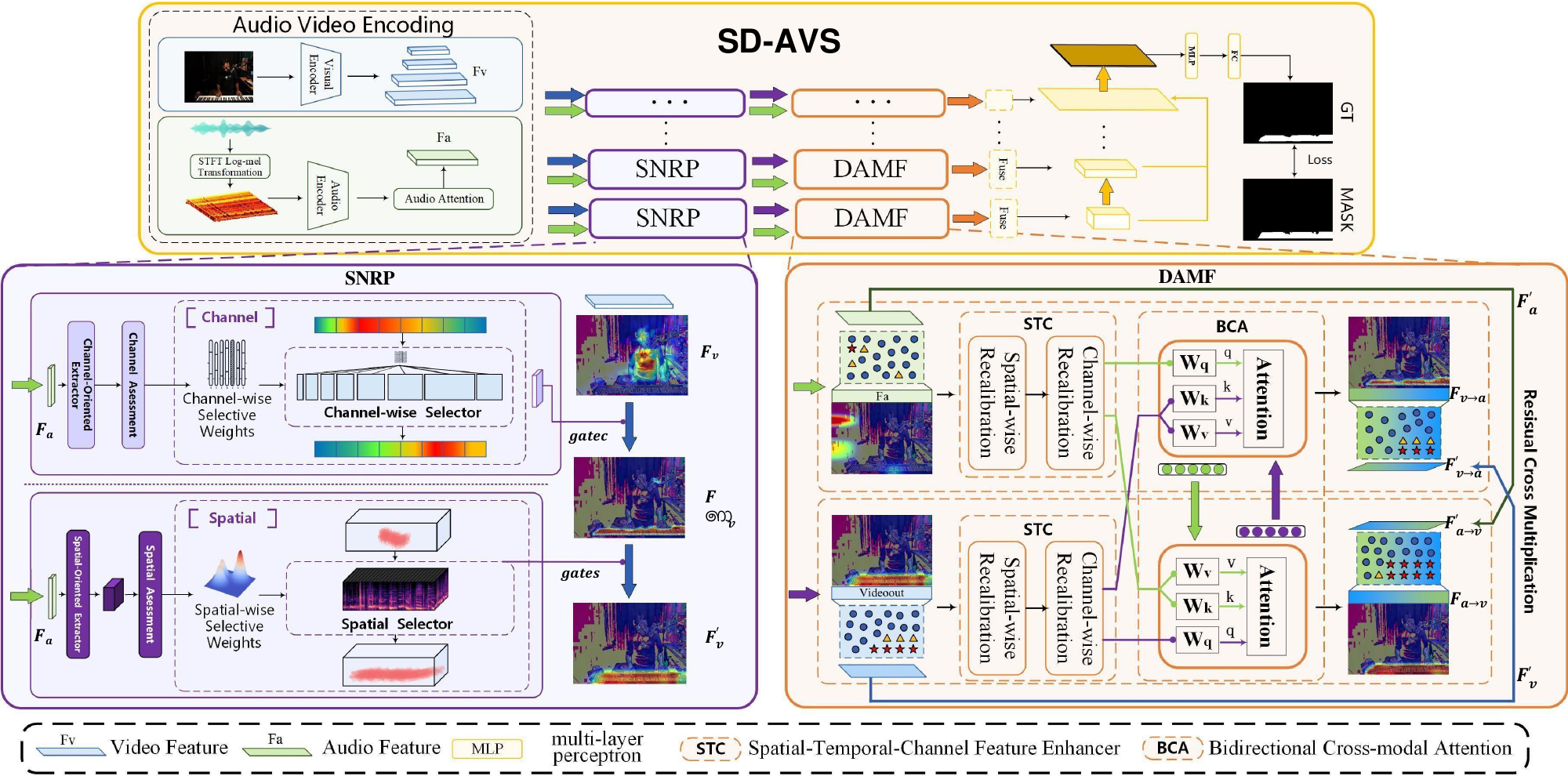}
  \caption{The overall pipeline of the proposed SDAVS, which includes the Selective Noise-Resilient Processor (SNRP) and Discriminative Audio-Visual Mutual Fusion (DAMF) modules. The SNRP module first filters noise and enhances audio-relevant features. The refined video features and original audio features are then fed into the DAMF module aligning their perception regions, which performs discriminative fusion using spatial-temporal-channel (STC) enhancement and bidirectional cross-modal attention. In DAMF, different icon shapes represent distinct perception regions per modality. After applying DAMF, the icons become more consistent, indicating improved cross-modal alignment.}
  \label{stucture}
\end{figure*}
\section{Method}
\subsection{Architecture Overview}
    We first describe the overall architecture of the SDAVS shown in Fig.~\ref{stucture}, which follows the encoder-decoder architecture. Following standard protocols~\cite{zhou2022audio}, we divide each $T$-second video into $T$ non-overlapping clips (typically 1 second per clip). We strictly maintain temporal alignment by extracting the visual frame and its corresponding audio segment from the same time interval, ensuring synchronized inputs for the subsequent network. In terms of the encoder, visual backbones take the video clips $V=[V_1,...V_t]$, $V_t\in R^{H \times W \times 3}$ as the input. It generates four video feature maps with different spatial resolution and channel number, as $\{F^i_v\}^4_{i=1} \in R^{B\times C^i\times T\times h^i\times w^i}$, where $(h^i, w^i)=(H, W)/2^{i+1}$. Meanwhile, the audio backbone utilizes the audio segment $A_t$ as input to generate audio features $F_a \in R^{B\times C\times T\times H\times W}$, where subscript $v$ represents video, subscript $a$ represents audio, $B$ is the batch size, $C$ is the number of channels, $T$ is the temporal dimension, and $H,W$ are the spatial dimensions.

The decoder structure consists of four progressive stages. The input of each stage consists of two features: $F_v$ and $F_a$. Each stage consists of two sequential modules: the Selective Noise-Resilient Processor (SNRP) followed by the Discriminative Audio-Visual Mutual Fusion (DAMF). First, $F_v$ and $F_a$ are fed into SNRP for noise suppression and key information enhancement. Subsequently, audio and video features are discriminatively fused and fully integrated across DAMF module, making them gradually consistent. Finally, the fused features are integrated with direct and skip connections in the decoder, thereby significantly enhancing the breadth and depth of mutual fusion. 
\subsection{Selective Noise-Resilient Processor}
The presence of noise interference and spectral overlap in audio signals poses significant challenges for precise audio-visual segmentation. This causes silent objects to be misinterpreted as sound-producing object, leading to the incorrect segmentation. Additionally, the structural differences between 1D audio signals and 2D video signals make their complete fusion difficult. To address these limitations, we propose the SNRP module, which can suppress interfered regions in audio signals and enhance the important regions.

Specifically, SNRP consists of the channel feature selector and spatial feature selector. It takes $F_v$ and $F_a$ in the form of a log-mel spectrogram~\cite{mukhamediya2023effect} derived from the Short-Time Fourier Transform (STFT)~\cite{mateo2018short} as input, decreasing the information loss caused by their different structure. Following the process, the SNRP module generates channel-wise and spatial-wise selective weights according to audio information. Low-weight selection suppresses noise and other audio interference, whereas high-weight selection enhances the key information, thereby improving the segmentation accuracy.

First, we utilize a channel feature selector mechanism to enhance audio and video features by adjusting the importance of each channel in the video feature map over dimensions, as:
\begin{equation}
\begin{gathered}
F'_a = Conv(UpSample(F_a)), \\
audio_{pool} = AdaptiveAvgPool(F'_a), \\
gatec =MLP(audio_{pool}),
\end{gathered}
\end{equation}
where $F'_a$ denotes the audio features after being upsampled and convolved to match the video features, $F'_a$ then go through global average pooling to gather spatial information in channel dimension, $audio_{pool}\in R^{B\times C\times T\times 1\times 1}$ is spatial-information squeezed audio features. The $MLP$ (multi-layer perceptron)~\cite{kruse2022multi} operation then outputs channel-wise selective weights $gatec$, which represents selective channel-wise weighting factors. The individual audio feature channels can be selectively regulated to the corresponding video feature map. Then, we exert $gatec$ on video feature maps. This step enhances spatial-aligned video feature regions and suppresses interfering regions:
\begin{equation}
\tilde F_v=F_v\odot gatec,
\end{equation}
where $\odot$ represents element-wise multiplication, $\tilde F_v$ represents adjusted video features. Important channel-wise information can be selectively enhanced, while irrelevant information is suppressed. Next, we apply an additional spatial feature selector mechanism to further refine audio and video features by adjusting the importance of each pixel in the video feature map. The spatial weighting parameters $gates$ are then applied on video features $\tilde F_v$ through element-wise multiplication:
\begin{equation}
\begin{gathered}
gates =\sigma( Conv({F'_a})),\\
F'_v=\tilde {F_v}\odot gates,
\end{gathered}
\end{equation}
where $gates$ is the discriminative spatial weights, $Conv$ is the convolutional operation, $\sigma$ is the $sigmoid$ function, and $F'_v$ is the output video features via SNRP. By utilizing the Sigmoid activation, the generated weights serve as a learnable soft-thresholding filter within the $[0, 1]$ range. During end-to-end training, these weights are optimized to automatically suppress irrelevant spatial areas (noise) while selectively highlighting relevant regions (sounding objects).
\subsection{Discriminative Audio-Visual Mutual Fusion}
Once the SNRP module effectively suppresses irrelevant audio noise and enhances key semantic cues, a critical challenge remains in achieving fine-grained and discriminative integration of audio-visual features. Previous methods like~\cite{senocak2023event} primarily focus on the guidance of audio in visual feature learning, fusing audio and visual features indiscriminately. Such undifferentiated fusion may lead to cross-modal interference, ultimately compromising the accuracy of sounding object localization and segmentation. While~\cite{chen2024bootstrapping} performs bidirectional cross-modal fusion, it focuses on addressing the imbalance between audio and visual modalities rather than mitigating audio interference and addressing inconsistent cross-modal perception issues. To address the challenge, we propose the DAMF module.

Given audio features $F'_a$ and video features $F'_v$, DAMF facilitates a precise and discriminative fusion through the Spatial-Temporal-Channel feature enhancer ($STC$) and bidirectional cross-modal interaction. First, explicitly addressing the complexity of cross-modal features, $STC$ is employed to enrich spatio-temporal representations while reducing computational cost via depthwise 3D convolutions~\cite{chollet2017xception}. Subsequently, the enhanced features are fused through the attention mechanism, formulated as:
\begin{equation}
\begin{gathered}
F_{a\rightarrow v} = Attn(STC_q(F'_v), STC_k(F'_a), STC_v(F'_a)),\\
F_{v\rightarrow a} = Attn(STC_q(F'_a), STC_k(F'_v), STC_v(F'_v)),
\end{gathered}
\end{equation}
where $F_{a\rightarrow v}$ and $F_{v\rightarrow a}$ denote the fused features using audio and video as queries, respectively. $Attn$ represents the standard attention mechanism. Specifically, the detailed implementation of $STC_x$ (where $x \in \{q,k,v\}$) is formulated as:
\begin{equation}
\begin{gathered}
\hat{F}'_{a/v} = Conv3d(F'_{a/v}), \\
STC_x = LN(CAR(\hat{F}'_{a/v})),
\end{gathered}
\end{equation}
where $Conv3d$ utilizes a kernel size of $(3, 3, 3)$ for query generation ($x=q$) and $(1,3,3)$ for key/value generation to reduce redundancy. $LN$ denotes Layer Normalization. To further refine feature discriminability, we introduce the Channel-wise Adaptive Recalibration ($CAR$) module:
\begin{equation}
    CAR(X) = X \odot \sigma(W_2 \cdot ReLU(W_1 \cdot GAP(X))),
\end{equation}
where $X$ is the input feature, $GAP$ denotes global average pooling, and $W_1, W_2$ are learnable weights for channel interaction. $\sigma$ and $\odot$ represent the Sigmoid function and element-wise multiplication, respectively.

The audio-video fusion and integration can be further improved by Residual Multiplication (RM), which emphasizes regions where audio and visual signals are mutually active, thereby reinforcing the alignment between modalities:
\begin{equation}
\begin{gathered}
F'_{a\rightarrow v} = F_{a\rightarrow v} \odot F'_a,\\
F'_{v\rightarrow a} = F_{v\rightarrow a} \odot F'_v,
\end{gathered}
\end{equation}
where $F'_{a\rightarrow v}$ and $F'_{v\rightarrow a}$ represent the result of RM on $F'_a$ and $F'_v$, thereby preserving original features as well as promoting the consistency of audio and video features.

In summary, the DAMF module enhances discriminative feature learning by establishing cross-modal correspondences in a bidirectional manner, allowing each modality to discriminatively attend to informative signals from the other. This mutual guidance mechanism significantly improves the model's ability to filter out irrelevant information, highlight semantically aligned patterns, and form more context-aware representations. Numerical and visual results have been demonstrated of the effectiveness of DAMF.
\begin{table*}
  \caption{Comparison with state-of-the-art methods on the S4 and MS3 settings. The best results are highlighted in bold, while the second-best results are underlined. $ \mathcal{J} \& \mathcal{F}_m $ means the average value of $\mathcal{J}$ and $\mathcal{F}_m$.}
  \setlength{\tabcolsep}{12pt}
  \renewcommand{\arraystretch}{1.01}
  \label{AVSexp}
  \centering
  \begin{tabular}{c c ccc ccc}
    \toprule
    \multirow{2}{*}{Method} & \multirow{2}{*}{Backbone} & \multicolumn{3}{c}{S4} & \multicolumn{3}{c}{MS3} \\
    \cmidrule(lr){3-5} \cmidrule(lr){6-8}
    & & $ \mathcal{J} $ &  $\mathcal{F}_m $ & $ \mathcal{J} \& \mathcal{F}_m $ & $ \mathcal{J} $ &  $\mathcal{F}_m $ & $ \mathcal{J} \& \mathcal{F}_m $\ \\
    \midrule
    TPAVI~\cite{zhou2022audio}& PVT-v2 & 78.7 & 87.9 & 83.3 & 54.0 & 64.5 & 59.3 \\
    AVSC~\cite{liu2023audio1} & Swin-base & 81.3 & 88.6 & 85.0 & 59.5 & 65.8 & 62.7 \\
    CATR~\cite{li2023catr}& PVT-v2 & 81.4 & 89.6 & 85.5 & 59.0 & 70.0 & 64.5 \\
    SelM~\cite{li2024selm}& PVT-v2 & 83.5 & 91.2 & 87.4 & 60.3 & 71.3 & 65.8 \\
    AQFormer~\cite{huang2023discovering}& PVT-v2 & 81.6 & 89.4 & 85.5 & 62.2 & 72.7 & 67.5 \\
    AVSAC~\cite{chen2024bootstrapping}& PVT-v2 & 84.5 & 91.6 & 88.1 & 64.2 & 76.6 & 70.4 \\
    AVSegFormer~\cite{gao2024avsegformer}& PVT-v2 & 83.1 & 90.5 & 86.8 & 61.3 & 73.0 & 67.2 \\
    AVSBG~\cite{hao2024improving}& PVT-v2 & 81.7 & 90.4 & 86.1 & 55.1 & 66.8 & 61.0 \\
    MUTR~\cite{yan2024referred}& Swin-Large & 81.5 & 89.8 & 85.7 & 65.0 & 73.0 & 69.0 \\
    GAVS~\cite{wang2024prompting} & ViT-base & 80.1 & 90.0 & 85.1 & 63.7 & 77.4 & 70.6 \\
    BAVS~\cite{liu2024bavs} & Swin-base & 82.7 & 89.8 & 86.3 & 59.6 & 65.9 & 61.3 \\
    QDFormer~\cite{li2024qdformer} & Swin-tiny & 79.5 & 88.2 & 83.9 & 61.9 & 66.1 & 64.0 \\
    COMBO~\cite{yang2024cooperation}& PVT-v2 & 84.7 & 91.9 & \underline{88.3} & 59.2 & 71.2 & 65.2 \\
    AVSStone~\cite{ma2024stepping} & Swin-base & 83.2 & 91.3 & 87.3 & 67.4 & 77.6 & 72.5 \\
    BiasAVS~\cite{sun2024unveiling} & Swin-base & 83.3 & \textbf{93.0} & 88.2 & 67.2 & \textbf{80.8} & \underline{74.0} \\
    C3N~\cite{10812843} & PVT-v2 & 83.1 & 90.8 & 86.95 & 61.7 & 72.2 & 66.95 \\
    PIF~\cite{10510606} & PVT-v2 & 81.4 & 90.0 & 85.7 & 58.9 & 70.9 & 64.9 \\
    CQFormer~\cite{10979212} & PVT-v2 & 83.6 & 91.2 & 87.4 & 61.0 & 72.7 & 66.9 \\
    AVS-Mamba~\cite{10891410} & PVT-v2 & \underline{85.0} & \underline{92.6} & \textbf{88.8} & \underline{68.6} & \underline{78.8} & 73.7 \\
    \midrule
    
    \rowcolor{red!5}
    SDAVS(Ours) & PVT-v2 & 84.6 & 91.3 & 88.0 & 70.5 & 79.0 & 74.8 \\
    \rowcolor{red!20}
    \textbf{SDAVS(Ours)} & Swin-base & \textbf{85.5} & 92.0 & \textbf{88.8} & \textbf{72.3} & \textbf{80.8} & \textbf{76.6} \\
    \bottomrule
  \end{tabular}
\end{table*}
\subsection{Multi-Scale Fusion Decoder with Skip Connections}
Our decoder employs a four-layer hierarchical architecture with multi-scale fusion to further enhance cross-modal discriminability. There are two ways to connect decoders: direct connection and skip connection: 
\begin{equation}
\begin{gathered}
\begin{aligned}
F^{j+1}_a, F^{j+1}_v = Fus_j\Bigl( &Conv\bigl(Up_j(F'_{a \rightarrow v} + F'_{v \rightarrow a},\\ 
& F^j_v)\bigr),Up_j(F^{j}_a) \Bigr),
\end{aligned} \\
Fus_{out} = \sum_{j=1}^{N} Up_{max}\bigl(Conv(F^{j}_a + F^{j}_v)\bigr),N = 4,
\end{gathered}
\end{equation}
where superscript $j$ denotes the decoder stage, $N$ denotes the number of decoder layers, $Fus_j$ denotes decoder operations. $Up$ dynamically adjusts the spatial dimensions of features to match $j+1$ scale of the next layer’s input. $Up_{max}$ refers to the upsampling based on the maximum resolution of the deepest layer in the network (final output layer). Direct connection $Up_j$ ensures continuity of contextual information in data transmission. Meanwhile, high semantic details from early layers are preserved through skip connection using $Up_{max}$, which upscales features to the maximum resolution of the final output layer, ensuring the retention of critical structural information. This design achieves complete fusion and gradual integration between audio features and video features. Finally, the model predicts the $Mask$ through a fully connected layer:
\begin{equation}
Mask = FC(MLP(Fus_{out})),
\end{equation}
where $FC$ means the fully connected layer. The output $Mask$ represents the predicted segmentation mask.
\section{Experiment}
\subsection{Experimental setup}
\subsubsection*{\bf Datasets}
To verify the effectiveness and rationality of the proposed model, we conduct experiments on three benchmark datasets: Semi-supervised Single-sound Source Segmentation (S4) and Fully-supervised Multiple-sound Source Segmentation (MS3)~\cite{zhou2022audio}, both of which are subsets of AVSBench-Object, as well as the Fully-supervised AVSBench-Semantic (AVSS) datasets~\cite{zhou2024audio}. S4 provides binary segmentation masks that identify the pixels corresponding to a single sounding object, with ground-truth available only for the initial frame during training. MS3 and AVSS, on the other hand, contain audio samples with multiple sounding objects and provide full supervision for every frame across the training videos. Compared with S4, the MS3 dataset involves distinguishing and segmenting multiple sound sources simultaneously, making it more challenging. AVSS further introduces semantic labels, requiring the model not only to localize sounding regions but also to recognize their semantic categories. Together, MS3 and AVSS present more challenging and realistic scenarios for audio-visual segmentation compared to the relatively simpler single-source S4.

\subsubsection*{\bf Training Details}
We conduct training and evaluation on these datasets using an NVIDIA A6000 GPU, with the visual backbone Pyramid Vision Transformer (PVT-v2)~\cite{wang2021pyramid} and Swin Transformer~\cite{liu2021swin}. We also use VGGish~\cite{hershey2017cnn} as the audio backbone to extract the audio feature from 96×64 log-Mel spectrograms. We employ AdamW as the optimizer and adopt a multi-step learning rate scheduler during training, with a batch size of 4 and an initial learning rate of $5\text{e}^{-5}$. Since the MS3 subset is quite small, we train it for 60 epochs, while the S4 and AVSS subsets are trained for 40 epochs. Consistent with previous works~\cite{gao2024avsegformer,gong2025avs}, we resize the input frames to 384×384. During training, our loss function includes cross entropy loss $L_{ce}$, IOU loss $L_{iou}$ and Dice loss $L_{dice}$. The final loss $L$ can be expressed as $L=L_{ce}+L_{iou}+L_{dice}$.
\subsubsection*{\bf Evaluation Metrics} For performance evaluation, we adopt two widely used metrics: Jaccard Index $ \mathcal{J} $ and F-score ($\mathcal{F}_m $), which quantify region-wise overlap and boundary alignment, respectively. In our experiments, we report the mean Jaccard Index ($ \mathcal{J} $) and mean F-score ($\mathcal{F}_m $) averaged across all test samples. For the F-score, following~\cite{zhou2022audio}, we set $m$ = 0.3 to prioritize precision over recall, reflecting the critical need to minimize erroneous activations in audio-visual segmentation tasks.
\begin{table}
  \caption{Comparison with state-of-the-art methods on the AVSS settings. The best results are highlighted in bold, while the second-best results are underlined. $ \mathcal{J} \& \mathcal{F}_m $ means the average value of $\mathcal{J}$ and $\mathcal{F}_m$. * denotes the two-stage methods.}
  \label{avssexp}
  \centering
  \setlength{\tabcolsep}{10pt}
  \begin{tabular}{c c ccc}
    \toprule
    \multirow{2}{*}{Method}      & \multirow{2}{*}{Backbone}    & \multicolumn{3}{c}{AVSS} \\
    \cmidrule(lr){3-5}
                &             & $ \mathcal{J} $     & $\mathcal{F}_m $    & $ \mathcal{J} \& \mathcal{F}_m $ \\
    \midrule
    TPAVI~\cite{zhou2022audio}& PVT-v2      & 29.8  & 35.2  & 32.5 \\
    CATR~\cite{li2023catr}    & PVT-v2      & 32.8  & 38.5  & 35.7 \\
    AVSegFormer~\cite{gao2024avsegformer}& PVT-v2      & 37.3  & 42.8  & 40.1 \\
    BAVS~\cite{liu2024bavs}    & Swin-base   & 33.6  & 37.5  & 35.6 \\
    SelM~\cite{li2024selm} & PVT-v2      & 41.3  & 46.9  & 44.1 \\
    AVSAC~\cite{chen2024bootstrapping} & PVT-v2      & 37.0  & 42.4  & 39.7 \\
    COMBO~\cite{yang2024cooperation}& PVT-v2      & 42.1  & 46.1  & 44.1 \\
    AVSStone*~\cite{ma2024stepping} & Swin-base  & \textbf{48.5} & \textbf{53.2} & \textbf{50.9} \\
    BiasAVS~\cite{sun2024unveiling} & Swin-base   & 44.4  & 49.9  & 47.2 \\
    CQFormer~\cite{10979212} & PVT-v2 & 38.1 & 41.0 & 39.55 \\
    AVS-Mamba~\cite{10891410} & PVT-v2 & 39.7 & 45.1 & 42.4 \\
    \midrule
    \rowcolor{gray!10}
    SDAVS(Ours)       & PVT-v2      & 44.8  & 50.2  & 47.5 \\
    \rowcolor{gray!30}
    \textbf{SDAVS(Ours)}  & Swin-base  & \underline{46.0}  & \underline{51.1}  & \underline{48.9} \\
    \bottomrule
  \end{tabular}
\end{table}
\begin{figure*}[ht]
  \centering
  \includegraphics[width=0.95\linewidth, height=0.5\linewidth]{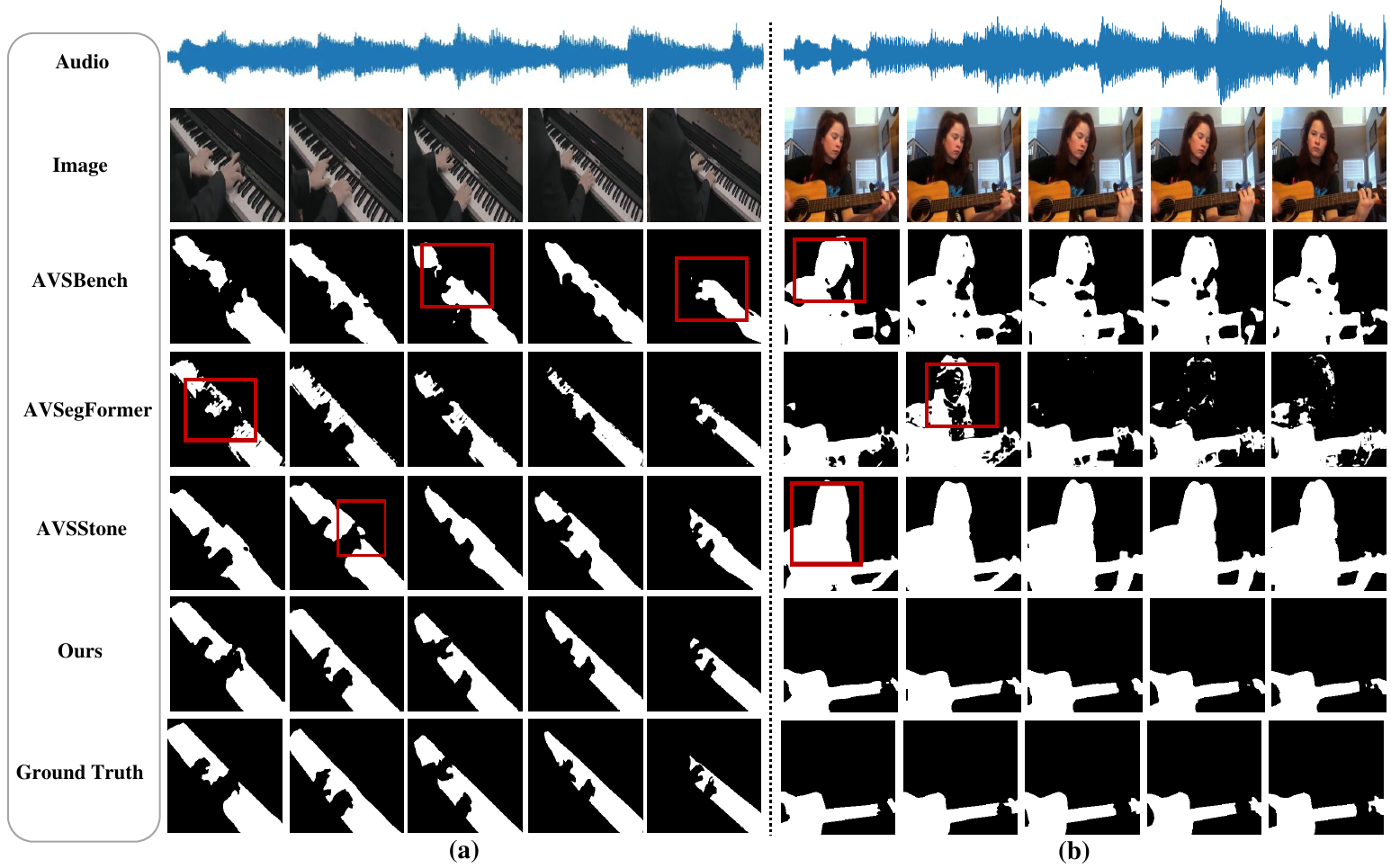}
  \caption{Qualitative comparison of AVSBench~\cite{zhou2022audio}, AVSegFormer~\cite{gao2024avsegformer}, AVSStone~\cite{ma2024stepping} and Ours on the AVSBench-object dataset. (a) demonstrates the accuracy of the model segmentation, and (b) demonstrates the model's suppression of the incorrect segmentation, i.e., the segmentation of non-sounding objects.}
  \label{avs}
\end{figure*}
\subsection{Comparisons with State-of-the-art methods}
We compare our proposed SDAVS against 19 recent state-of-the-art AVS methods on the S4, MS3, and AVSS datasets. To ensure a fair comparison, we follow standard protocols by employing a transformer-based backbone for visual feature extraction and VGGish for audio feature extraction.
\subsubsection*{\bf Quantitative Evaluation} Table~\ref{AVSexp} shows the quantitative comparisons between our model and other recent methods on the S4 and MS3 datasets. We can see that our model significantly outperforms these methods. Specifically, our model achieves a significant improvement on $ \mathcal{J} \& \mathcal{F}_m $, with gains of 0.6 on the S4 datasets and 2.6 on the MS3 datasets, highlighting its robustness across both single-source and multi-source segmentation tasks. Notably, the improvement in the MS3 datasets is significantly higher than the improvement in the S4 datasets. We attribute to the issue that our method extracts key information in the audio while suppressing noise interference for effective audio-video fusion, which can cope especially with the complexity of the MS3 task. Additionally, Table~\ref{avssexp} offers a detailed comparison of our method with others on the AVSS datasets. Similarly, our model improves an impressive 1.7 improvement on $ \mathcal{J} \& \mathcal{F}_m $ over the previous one-stage state-of-the-art. The competitive performance underscores the effectiveness of our noise-resilient and discriminative mutual interaction strategies.
\begin{figure}[t]
  \centering
  \includegraphics[width=\linewidth]{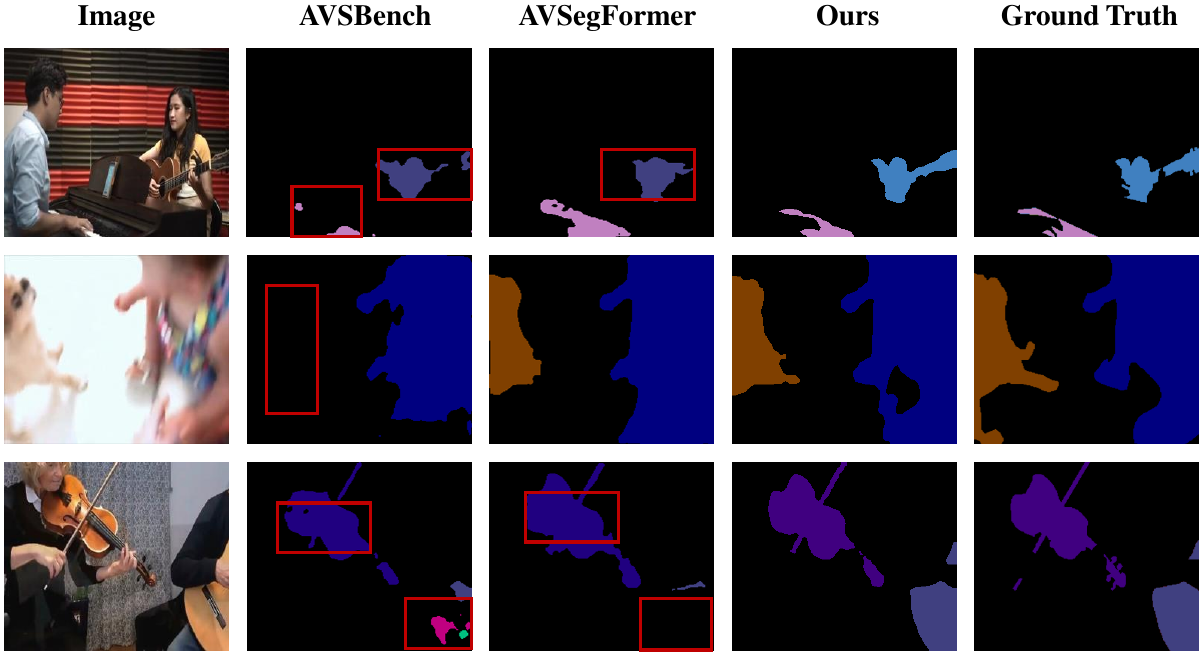}
  \caption{Qualitative comparison between AVSBench~\cite{zhou2022audio}, AVSegFormer ~\cite{gao2024avsegformer} and Ours on the AVSS dataset, demonstrating our model's excellent ability to interpret semantic information.}
  \label{avss}
\end{figure}
\subsubsection*{\bf Qualitative Evaluation} To showcase the visual effect of our model’s segmentation effects, we display the comparative visual results from AVSBench~\cite{zhou2022audio}, AVSegFormer~\cite{gao2024avsegformer}, and ours. Fig.~\ref{avs} presents the result on S4 and MS3 settings. It can be found that our segmentation results show more fine details compared with other methods. As shown in Fig.~\ref{avs}(a), the piano and hands can not be segmented well by previous methods, especially in the details of the edges. Additionally, Fig.~\ref{avs}(b) presents a woman playing guitar. It can be observed that previous methods suffer from incorrect segmentation of sounding objects. However, our model demonstrates stronger sound source localization and avoids segmentation of irrelevant objects. Meanwhile, Fig.~\ref{avss} shows the AVSS settings, and the results demonstrate our model has strong semantic understanding and is able to effectively recognize the correct sound source in complex scene.
\begin{table}[t]
  \caption{The impact of the proposed modules on the model's effectiveness, and the absence of any module leads to performance degradation.}
  \label{moduleabliation}
  \centering
  \begin{tabular}{lcccccc}
    \toprule
    \multirow{2}{*}{Module}    & \multicolumn{3}{c}{S4} & \multicolumn{3}{c}{MS3} \\
    \cmidrule(lr){2-4} \cmidrule(lr){5-7}
             & $ \mathcal{J} $    & $\mathcal{F}_m $    & $ \mathcal{J} \& \mathcal{F}_m $ & $ \mathcal{J} $    & $\mathcal{F}_m $    & $ \mathcal{J} \& \mathcal{F}_m $\\
    \midrule
    w/o SNRP & 84.1  & 90.8  & 87.5  & 70.7  & 79.0  & 74.9  \\
    w/o DAMF & 83.2  & 90.5  & 86.9  & 69.1  & 78.2  & 73.7  \\
    \textbf{SDAVS(Ours)}     & \textbf{85.5} & \textbf{92.0} & \textbf{88.8} & \textbf{72.3} & \textbf{80.8} & \textbf{76.6} \\
    \bottomrule
  \end{tabular}
\end{table}
\begin{table}[t]
  \caption{Ablation study on SNRP components (CFS and SFS).}
  \label{SNRPabla}
  \centering
  \begin{tabular}{lcccccc}
    \toprule
    \multirow{2}{*}{SNRP} & \multicolumn{3}{c}{S4} & \multicolumn{3}{c}{MS3} \\
    \cmidrule(lr){2-4} \cmidrule(lr){5-7}
          & $ \mathcal{J} $    & $\mathcal{F}_m $    & $ \mathcal{J} \& \mathcal{F}_m $ & $ \mathcal{J} $    & $\mathcal{F}_m $    & $ \mathcal{J} \& \mathcal{F}_m $\\
    \midrule
    w/o CFS & 84.3  & 90.9  & 87.6  & 70.6  & 79.1  & 74.9  \\
    w/o SFS & 84.6  & 91.3  & 88.0  & 70.5  & 79.1  & 74.8  \\
    \textbf{SDAVS(Ours)}    & \textbf{85.5}  & \textbf{92.0}  & \textbf{88.8}  & \textbf{72.3}  & \textbf{80.8}  & \textbf{76.6}  \\
    \bottomrule
  \end{tabular}
\end{table}
\begin{figure}[t]
  \centering
  \includegraphics[width=\linewidth, height=0.55\linewidth]{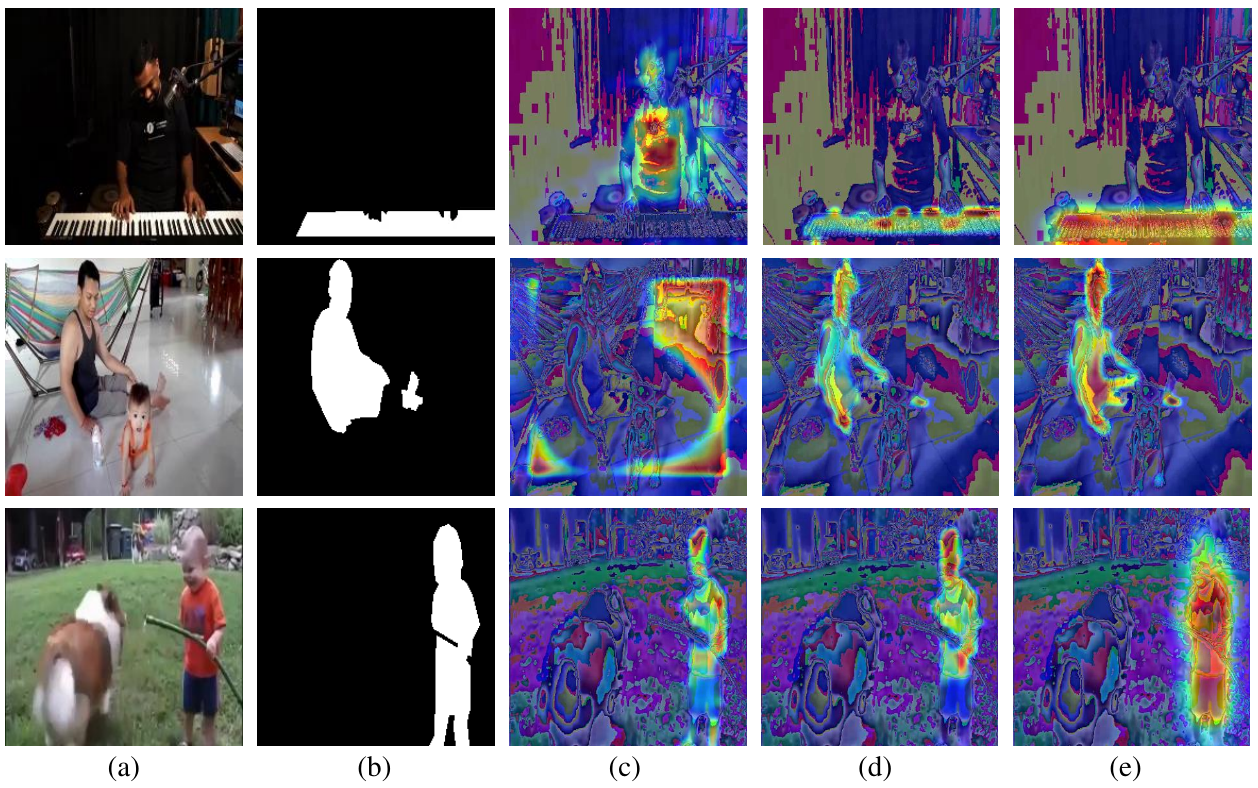}
  \caption{feature maps visualizations in SNRP presented in sequence from left to right. (a) raw images, (b) ground truth masks, (c) feature maps before processing by SNRP, (d) feature maps after processing CFS, and (e) feature maps after processing SFS.}
  \label{SNRP}
\end{figure}
\subsection{Ablation Experiments}
In order to achieve a representative analysis, we perform ablation studies in S4 and MS3 datasets, and all experiments used the Swin-Transformer backbone as the visual encoder.

\subsubsection*{\bf Overall Module Analysis} 
To evaluate the contribution of SNRP and DAMF to the overall model performance, we conduct an ablation study by removing each module individually from the full architecture. As presented in Table~\ref{moduleabliation}, removing DAMF module leads to a noticeable performance drop, with a $ \mathcal{J} \& \mathcal{F}_m $ decrease of 1.9 on S4 and 2.9 on MS3. On the other hand, the exclusion of SNRP also degrades performance, resulting in a $ \mathcal{J} \& \mathcal{F}_m $ decrease of 1.3 on S4 and 1.7 on MS3. In particular, we can see that the exclusion of DAMF leads to a more substantial decline. These results quantitatively confirm that both SNRP and DAMF play complementary and indispensable roles in improving the model’s ability to perform fine-grained, precise segmentation results.
\subsubsection*{\bf Effects of Noise-Resilient strategies in SNRP} 
In order to investigate the internal contributions of SNRP, we conduct additional ablation studies by removing the semantic channel feature selector (CFS) and spatial feature selector (SFS) in SNRP respectively. As demonstrated in Table~\ref{SNRPabla}, The removal of SFS and CFS individually results in notable performance degradation, numerically proving both of them are indispensable in SNRP. Notably, our results are more pronounced on the MS3 dataset, where the removal of CFS and SFS leads to a decrease of about 1.7 and 1.8 on $ \mathcal{J} \& \mathcal{F}_m $, respectively, compared to smaller reductions of 1.2 and 0.8 on the S4 dataset. These results suggest that the contributions of both selectors are particularly critical in more complex multi-source scenarios.

Furthermore, to prove the effects of Noise-Resilient strategies from a more intuitive perspective, we employ a qualitative analysis approach and generate feature maps at different stages of SNRP. Meanwhile, we employed Grad-CAM, a gradient-based visualization technique, to analyze the module's impact on semantic perception and attention by comparing feature maps, as shown in Fig.~\ref{SNRP}. Specifically, we visualize the feature map before it is input into the SNRP (denoted as Fig.~\ref{SNRP}(c)), as well as the feature maps output from CFS (denotes as Fig.~\ref{SNRP}(d)) and the SFS (denotes as Fig.~\ref{SNRP}(e)), shown in Fig.~\ref{SNRP}. It can be observed from the visual result that the model progressively enhances its focus on the sounding object as features are processed through the two selectors in the SNRP module. Specifically, in the first stage, the feature map Fig.~\ref{SNRP}(d) gradually directs attention toward the sound-producing object, compared to the feature map Fig.~\ref{SNRP}(c). In the second stage, following the processing by SFS, the feature map Fig.~\ref{SNRP}(e) becomes even more focused on the sounding object, while the model effectively suppresses attention to irrelevant background noise, thereby improving the overall segmentation precision. This indicates a strengthened activation in the target region, accompanied by an effective suppression of background interference, thereby demonstrating the module’s capability to shift attention towards sound-relevant visual cues. To further validate the effectiveness of SNRP in blocking noise propagation, we evaluated a variant where the module is placed after the fusion stage, as shown in Table~\ref{tab:R1}. The results confirm that SNRP is crucial for source filtering, as post-fusion placement fails to rectify noise propagated during interaction.
\begin{table}
  \caption{Ablation study on the placement of the SNRP module. 'Pre' indicates placing SNRP before the fusion stage, while 'After' indicates placing it after the fusion stage.}
  \label{tab:R1}
  \centering
  \begin{tabular}{lcccccc}
    \toprule
    \multirow{2}{*}{Position} & \multicolumn{3}{c}{S4} & \multicolumn{3}{c}{MS3} \\
    \cmidrule(lr){2-4} \cmidrule(lr){5-7}
           & $ \mathcal{J} $    & $\mathcal{F}_m $    & $ \mathcal{J} \& \mathcal{F}_m $ & $ \mathcal{J} $    & $\mathcal{F}_m $    & $ \mathcal{J} \& \mathcal{F}_m $  \\
    \midrule
    Pre   & \textbf{85.5} & \textbf{92}  & \textbf{88.8} & \textbf{72.3} & \textbf{80.8} & \textbf{76.6} \\
    After & 84.3 & 90.6 & 87.5 & 71.1 & 78.7 & 74.9 \\
    \bottomrule
  \end{tabular}
\end{table}

\begin{table}
  \caption{Ablation study on DAMF. A$\to$V means audio-prompt method and V$\to$A means video-prompt method.}
  \label{DAMFabla}
  \centering
  \begin{tabular}{lcccccc}
    \toprule
    \multirow{2}{*}{DAMF}      & \multicolumn{3}{c}{S4} & \multicolumn{3}{c}{MS3} \\
    \cmidrule(lr){2-4} \cmidrule(lr){5-7}
              & $ \mathcal{J} $    & $\mathcal{F}_m $    & $ \mathcal{J} \& \mathcal{F}_m $ & $ \mathcal{J} $    & $\mathcal{F}_m $    & $ \mathcal{J} \& \mathcal{F}_m $\\
    \midrule
    only A$\to$V & 83.4  & 90.9  & 87.2  & 69.6  & 78.8  & 74.2  \\
    only V$\to$A & 83.2  & 90.7  & 87.0  & 69.3  & 78.6  & 74.0  \\
    w/o STC     & 83.8  & 91.0  & 87.4  & 70.5  & 79.3  & 74.9  \\
    \midrule
    \textbf{SDAVS(Ours)}        & \textbf{85.5} & \textbf{92.0} & \textbf{88.8} & \textbf{72.3} & \textbf{80.8} & \textbf{76.6} \\
    \bottomrule
  \end{tabular}
\end{table}

\noindent\textbf{Effects of fusion strategies in DAMF.} 
To evaluate the contribution of each component within the DAMF module, we conduct ablation experiments focusing on STC and the bidirectional interaction mechanism in DAMF. As shown in Table~\ref{DAMFabla}, “only a$\to$v” denotes the audio-prompted variant and “only v$\to$a” denotes the video-prompted variant. We can see in Table~\ref{DAMFabla} that removing STC or replacing the bidirectional design with a unidirectional variant leads to notable performance degradation on both the S4 and MS3 datasets. Similarly, the degradation is more substantial on the MS3 dataset. These results numerically confirm that both STC and the bidirectional interaction scheme are indispensable for maximizing the module's cross-modal fusion capabilities.

\begin{figure}[t]
  \centering
  \includegraphics[width=\linewidth]{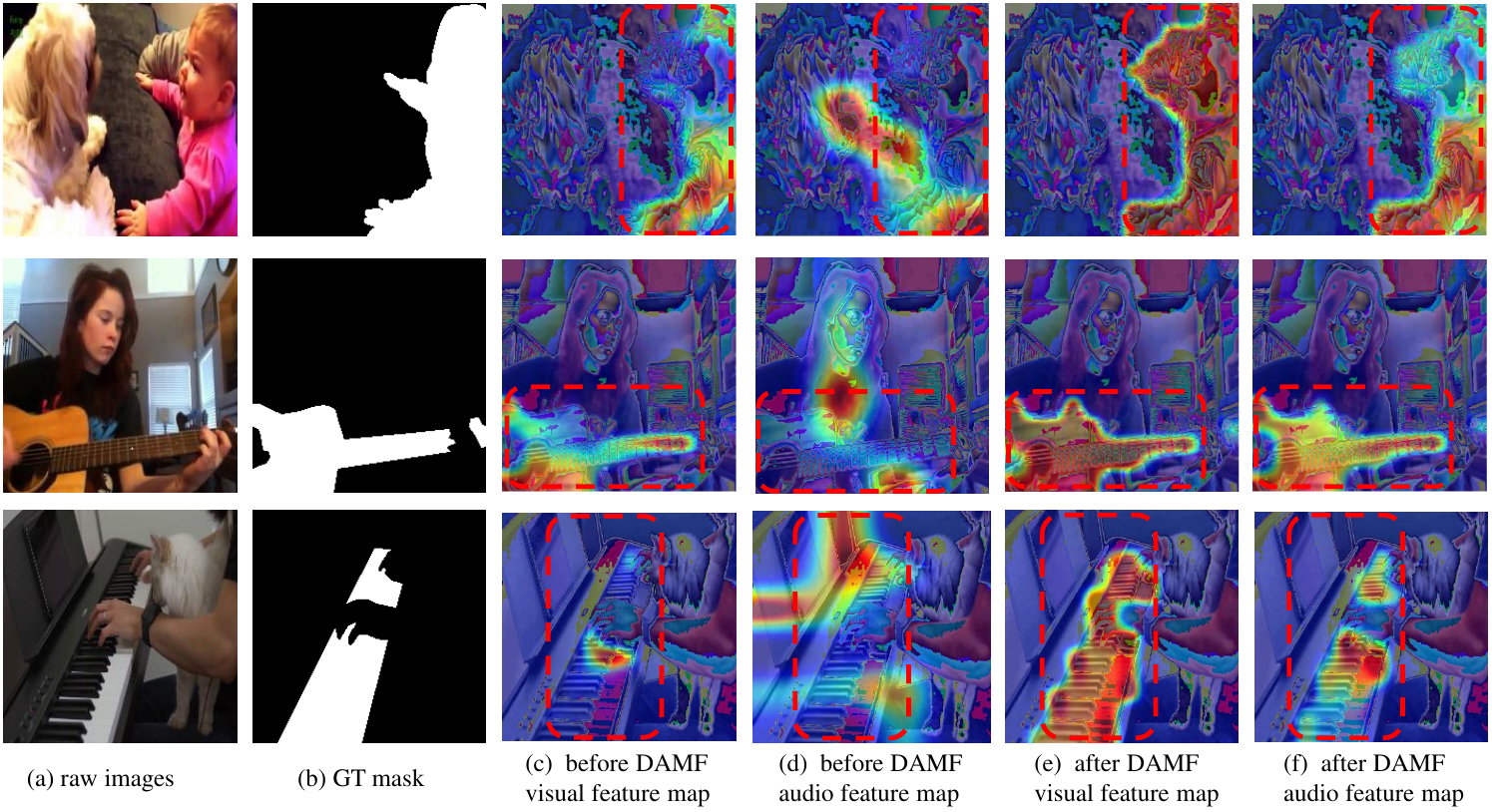}
  \caption{feature maps visualizations in DAMF presented in sequence from left to right. (a) raw images, (b) ground truth masks, (c) visual feature maps before processing by DAMF, (d) audio feature maps before processing by DAMF, (e) visual feature maps after processing by DAMF, and (f) audio feature maps after processing by DAMF. The red boxes highlight the primary regions of modal attention, illustrating the significant spatial discrepancy before fusion and the improved cross-modal consensus after DAMF processing.}
  \label{DAMF}
\end{figure}

Moreover, to provide a more perceptual perspective for evaluating the discriminative mutual interaction in DAMF, we visualize the visual and audio feature maps before and after implementing DAMF, respectively. As shown in Fig.~\ref{DAMF}, prior to the DAMF module, we can see that video-prompt feature map (denotes as Fig.~\ref{DAMF}(c)) and audio-prompt feature map (denotes as Fig.~\ref{DAMF}(d)) demonstrate significant spatial discrepancies, reflecting modality-specific focus and a lack of consistency. However, after processing through the DAMF module, the video feature map (denotes as Fig.~\ref{DAMF}(e)) and the audio feature map (denotes as Fig.~\ref{DAMF}(f)) exhibit enhanced consistency, indicating that the module effectively facilitates discriminative mutual integration. This consistent attention highlights the module’s capacity to jointly uncover shared salient patterns across modalities while selectively emphasizing correlated components and suppressing irrelevant or uncorrelated information. To quantitatively evaluate the improvement in perception consistency facilitated by DAMF, we employed common metrics including KL Divergence, CKA, and JS Divergence. As shown in Table~\ref{tab:metrics_comparison}, the significant improvement in consistency mathematically validates the effectiveness of the DAMF module.
\begin{table}[!t]
  \caption{Quantitative analysis of cross-modal perception consistency before and after the DAMF module. CKA (Centered Kernel Alignment), KL Divergence, and JS Divergence are employed to measure the feature distribution alignment.}
  \label{tab:metrics_comparison}
  \centering
  \setlength{\tabcolsep}{12pt}
  \begin{tabular}{lccc}
    \toprule
    Processing Stage  & CKA$\uparrow$ & KL$\downarrow$ & JS$\downarrow$ \\
    \midrule
    before DAMF &  0.33 &  22.1 &  0.43 \\
    After DAMF  & \textbf{0.79} & \textbf{15.3} & \textbf{0.32} \\
    \bottomrule
  \end{tabular}
  \vspace{-5pt}
\end{table}

\noindent\textbf{Internal Design Validation of DAMF.} 
Beyond the overall module effectiveness, we further validate the specific design choices within DAMF, focusing on the cross-modal interaction strategy and the necessity of sub-modules. Regarding the interaction strategy, the design objective of our Residual Multiplication (RM) is to achieve "Spatial Consensus" between modalities. Unlike addition, which tends to accumulate disparate information, element-wise multiplication emphasizes regions where audio and visual signals are mutually active, thereby reinforcing the alignment between modalities. We validate this design against variants (e.g., straight connection and addition) in Table~\ref{tab:interaction_ablation}. The results confirm that multiplication outperforms addition, proving that enforcing consensus is superior to simple aggregation.

\begin{figure}[!t]
  \centering
  \includegraphics[width=\linewidth]{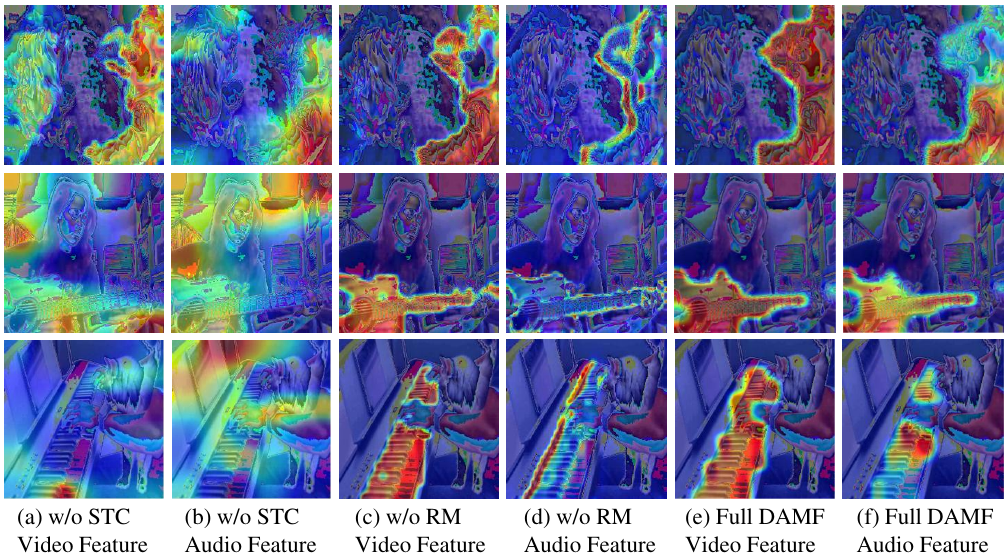}
  \caption{Visualization comparisons of feature maps under different ablation settings within the DAMF module. The columns from left to right display: (a-b) Video and Audio features without the STC module; (c-d) Video and Audio features without Residual Multiplication; and (e-f) Video and Audio features processed by the full DAMF module.}
  \label{R2}
\end{figure}

To intuitively verify the contributions of each component within DAMF, we provide a qualitative ablation comparison in Fig.~\ref{R2}. As shown, removing the STC module (columns a, b) leads to significant background noise interference, while removing Residual Multiplication (columns c, d) results in ambiguous object boundaries. These observations explicitly highlight our design rationale, confirming the complementary roles of STC in feature enhancement and Residual Multiplication in cross-modal interaction.

\begin{table}
  \caption{Performance comparison of different cross-modal interaction strategies within the DAMF module. Three strategies including straight-through connection, element-wise addition, and the proposed residual multiplication ($\odot$) are evaluated on the S4 and MS3 datasets.}
  \label{tab:interaction_ablation}
  \centering
  \begin{tabular}{lcccccc}
    \toprule
    \multirow{2}{*}{Interaction} & \multicolumn{3}{c}{S4} & \multicolumn{3}{c}{MS3} \\
    \cmidrule(lr){2-4} \cmidrule(lr){5-7}
             & $\mathcal{J}$ & $\mathcal{F}_m$ & $ \mathcal{J} \& \mathcal{F}_m $ & $\mathcal{J}$ & $\mathcal{F}_m$ & $ \mathcal{J} \& \mathcal{F}_m $ \\
    \midrule
    straight & 83.9 & 91.2 & 87.6 & 71.2 & 79.6 & 74.8 \\
    add      & 84.3 & 91.5 & 87.9 & 71.6 & 79.9 & 75.2 \\
    $\odot$  & \textbf{85.5} & \textbf{92.0} & \textbf{88.8} & \textbf{72.3} & \textbf{80.8} & \textbf{76.6} \\
    \bottomrule
  \end{tabular}
\end{table}

\subsection{Analysis of Noise Impact}

To quantitatively assess the impact of audio interference on model performance and to verify the effectiveness of the proposed SNRP module, we conduct an experiment to evaluate the model’s robustness against audio interference. Specifically, we employ an additive mixing strategy to construct the noisy evaluation set. We generate noise sequences that match the exact duration of the original audio clips and superimpose them onto the entire temporal span with a scaling factor of 0.1, thereby simulating continuous global interference. Regarding the noise types, we select Brownian noise and ``moving train'' sounds for their representative characteristics. We choose Brownian noise over standard white or pink noise because its stronger low-frequency energy closely mimics common environmental background sounds, such as wind and machinery, providing a more realistic assessment of robustness. Furthermore, the ``moving train'' noise is included to simulate real-world traffic interference, introducing complex acoustic patterns that challenge the model's generalization capabilities beyond stationary background noise. We compare the model performance with and without the SNRP module under these conditions. Table~\ref{noiseinterference} proves that SNRP significantly enhances robustness to audio interference, yielding improved performance under both synthetic and real-world noise. In particular, the performance degradation caused by either Brownian noise or real-world train noise is marginal and can be considered negligible when the SNRP module is employed, while removing SNRP module leads to a noticeable performance drop on both conditions, with a $ \mathcal{J} \& \mathcal{F}_m $ decrease of 2.2 and 1.5, respectively.

\subsection{Statistical Justification of SNRP}

\begin{figure}[t]
  \centering
  \includegraphics[width=0.9\linewidth , height=0.8\linewidth]{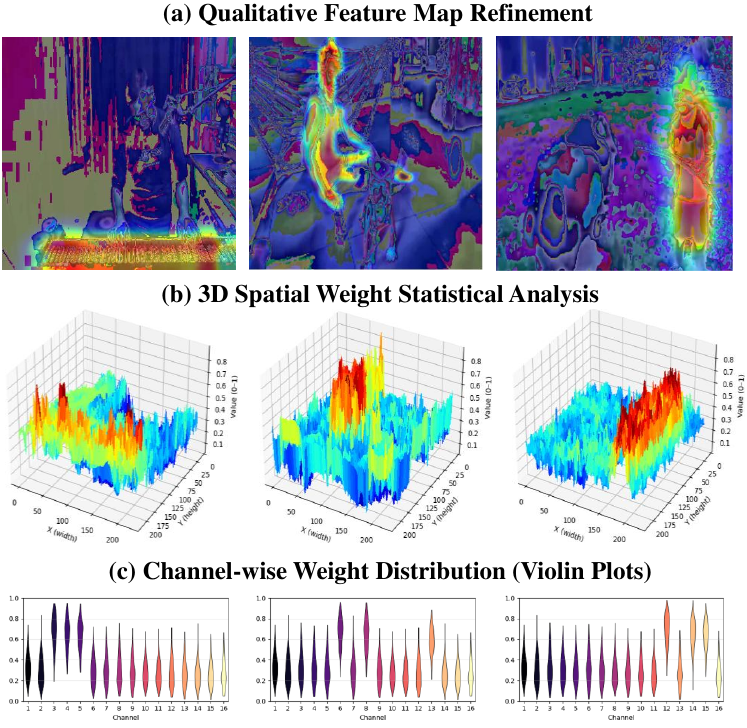}
  \caption{Statistical analysis of signal-noise separation in SNRP. (a) Qualitative Feature Map Refinement: Feature maps from the final stage of SNRP (Ref. Fig. 5), showing refined spatial focus. (b) 3D Spatial Weight Statistical Analysis: 3D response maps revealing the flattening of background noise into a low-response floor and sharpening of signal peaks. (c) Channel-wise Weight Distribution: Violin plots of channel weights demonstrating the heterogeneous distribution and selective channel-wise gating.}
  \label{R1}
\end{figure}

To statistically validate the signal-noise separation capability of SNRP, we visualize the feature response distribution in spatial and channel dimensions, as shown in Fig.~\ref{R1}. Spatially, the 3D response maps in Fig.~\ref{R1}(b) provide direct evidence of spatial noise suppression. The clear contrast between the suppressed background and the sharp target peaks confirms that the module effectively separates the sounding object from background noise. At the channel level, the weight distribution analysis in Fig.~\ref{R1}(c) validates the selection ability of our mechanism. Instead of treating all channels equally, the varied distribution indicates that SNRP successfully distinguishes between channels, actively filtering out noisy channels while preserving those containing useful information.
\begin{table}[!t]
\caption{Robustness evaluation of SDAVS with and without the SNRP module under different audio interference conditions. This table presents the segmentation performance on the MS3 dataset under three audio conditions: clean (original) audio, synthetic Brownian noise, and real-world train noise. }
\label{noiseinterference}
\centering
\footnotesize
\begin{tabular}{l ccc ccc}
\toprule
\multirow{2}{*}{Test Condition} & \multicolumn{3}{c }{w/o SNRP} & \multicolumn{3}{c}{Ours} \\
\cmidrule(r){2-4} \cmidrule(l){5-7}
 & $\mathcal{J}$ & $\mathcal{F}_m$ & $ \mathcal{J} \& \mathcal{F}_m $ & $\mathcal{J}$ & $\mathcal{F}_m$ & $ \mathcal{J} \& \mathcal{F}_m $ \\
\midrule
Original Audio & 70.7 & 79.0 & 74.9 & \textbf{72.3} & \textbf{80.8} & \textbf{76.6} \\
Brownian Noise & 68.8 & 76.5 & 72.7 & 72.1 & 80.5 & 76.3 \\
Moving Train   & 69.5 & 77.3 & 73.4 & 72.2 & 80.8 & 76.5 \\
\bottomrule
\end{tabular}
\end{table}

\subsection{Computational performance Analysis}
\begin{figure}[t]
  \centering
  \includegraphics[width=\linewidth , height=0.85\linewidth]{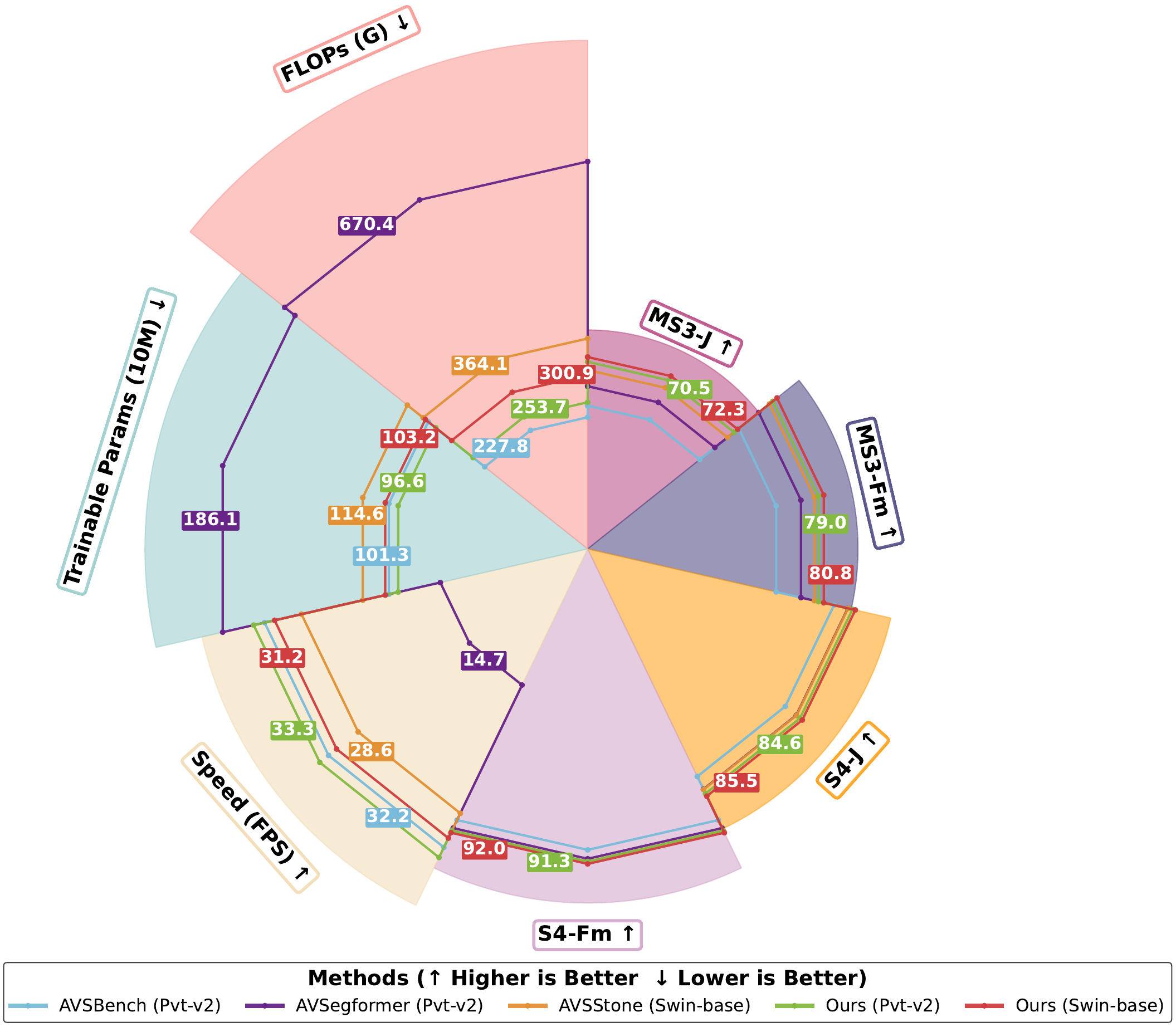}
  \caption{Comparison of model efficiency across state-of-the-art methods in terms of trainable parameters, computational complexity (FLOPs), and speed. Speed refers to the number of frames processed per second during inference in the same experimental settings.}
  \label{complex}
\end{figure}

To evaluate the efficiency of our proposed SDAVS, we compare it with several state-of-the-art audio-visual segmentation models in terms of model complexity and inference performance. As illustrated in Fig.~\ref{complex}, our method achieves competitive efficiency across both PVT-v2 and Swin-base backbones. Specifically, compared to AVSegformer, our model with PVT-v2 reduces the trainable parameters by nearly 48\% and lowers FLOPs by over 60\%, while more than doubling the inference speed (32.25 vs. 14.71 frames per second). Even when compared to other efficient frameworks like AVSBench~\cite{zhou2022audio} and AVSStone~\cite{ma2024stepping}, our method maintains a comparable or better inference speed, with significantly fewer parameters and moderate computational cost. These results clearly demonstrate the practicality of SDAVS for real-time or resource-constrained scenarios, validating its architectural efficiency without sacrificing segmentation quality.

To provide a transparent view of the model's complexity, we conduct a detailed component-wise efficiency analysis in Table~\ref{tab:complexity_comparison}. Regarding the macro-modules, the SNRP module is proven to be extremely lightweight, adding negligible FLOPs (300.85 G vs. 300.9 G) and having virtually no impact on inference speed. Regarding the internal sub-modules within DAMF, the STC module is critical for efficiency. As shown in the table, removing STC (i.e., operating on high-dimensional features) dramatically increases computational cost to 426.7 G FLOPs and drops speed to 22 FPS. This confirms that STC effectively reduces redundancy. Finally, the Residual Multiplication (RM) adds no extra parameters or computational burden, ensuring efficient cross-modal interaction.
\begin{table}
  \caption{Component-wise computational efficiency analysis. We report FLOPs, Parameters, and Inference Speed (FPS) for the removal of SNRP, DAMF, and internal sub-modules (STC, RM) to comprehensively validate the architectural trade-offs between complexity and performance.}
  \label{tab:complexity_comparison}
  \centering
  \begin{tabular}{lccc}
    \toprule
    Method   & FLOPs (G)$\downarrow$ & Trainable Params (M)$\downarrow$ & FPS$\uparrow$  \\
    \midrule
    w/o SNRP & 300.85    & 100.58               & 31.1 \\
    w/o DAMF & 280.31    & 100.76               & 33.5 \\
    w/o STC  & 426.7     & 103                  & 22   \\
    w/o RM   & 300.88    & 103.2                & 31.2 \\
    Ours     & 300.9     & 103.2                & 31.2 \\
    \bottomrule
  \end{tabular}
\end{table}
\begin{table}[!t]
\caption{Performance comparison of Ours on the AVIS dataset}\label{AVIS}
\centering
\footnotesize
\setlength{\tabcolsep}{12pt}
\begin{tabular}{lccc}
\toprule
Method        & FSLA & HOTA & mAP \\
\midrule
AVSegformer~\cite{gao2024avsegformer}   & 35.7 & 55.7 & 35.7 \\
COMBO~\cite{yang2024cooperation}         & 39.5 & 57.4 & 37.5 \\
Ours          & \underline{41.5} & \underline{60.3} & \underline{39.8} \\
AVISM~\cite{guo2025audio}         & \textbf{42.8} & \textbf{61.7} & \textbf{40.6} \\
\bottomrule
\end{tabular}
\end{table}
\subsection{Generalization on AVIS and VPO Benchmarks}
To assess the generalizability of our model beyond conventional AVS settings, we conduct additional experiments on the AVIS dataset~\cite{guo2025audio}, which comprises 926 videos averaging 61.4 seconds in length and covering diverse audio categories such as music, speaking, and ambient sounds. Notably, we apply our model directly to AVIS without any fine-tuning or retraining. As shown in Table~\ref{AVIS}, our method outperforms previous AVS approaches in all evaluation metrics (FSLA, HOTA, and mAP), such as AVSegformer~\cite{gao2024avsegformer}, COMBO~\cite{yang2024cooperation}. Furthermore, our method achieves comparable performance with the AVISM model specifically designed for AVIS, demonstrating that the proposed framework possesses strong generalizability to longer and more diverse video sequences and maintains robustness under varying audio conditions. This highlights the model’s potential applicability to broader real-world scenarios beyond standard AVS benchmarks.

To make our generalization evaluation more rigorous, we also extend experiments to the Visual Post-production (VPO) benchmark. Unlike standard datasets, VPO is constructed by matching COCO images with VGGSound audio based on semantic consistency, effectively mitigating the 'common-sense bias' often found in existing benchmarks. It includes three settings: single-source (SS), multi-source (MS), and the highly challenging multi-source multi-instance (MSMI). As presented in Table~\ref{tab:vpo_comparison_single}, we compare SDAVS with the state-of-the-art method CAVP (the proposer of VPO). Benefiting from the noise suppression of SNRP and the spatial consensus of DAMF, our model demonstrates exceptional robustness on these unbiased benchmarks. Notably, under the most difficult VPO-MSMI setting—which demands discriminating multiple instances of the same category via spatial audio—SDAVS outperforms CAVP by +2.3\% in mIoU. This consistent superiority confirms that our method effectively generalizes to complex, realistic, and unbiased scenarios beyond standard AVS tasks.
\begin{table}[!t]
  \caption{Comparison results on VPO-SS, VPO-MS, and VPO-MSMI datasets.}
  \label{tab:vpo_comparison_single}
  \centering
  \footnotesize
  
  \setlength{\tabcolsep}{14pt} 
  
  \begin{tabular}{cccc}
    \toprule
    Method & mIoU $\uparrow$ & F-score $\uparrow$ & FDR $\downarrow$ \\
    \midrule
    \multicolumn{4}{c}{\textbf{\textit{Dataset: VPO-SS}}} \\
    \midrule
    TPAVI~\cite{zhou2022audio}    & 52.75 & 69.54 & 22.83 \\
    AVSegformer~\cite{gao2024avsegformer} & 57.55 & 73.03 & 19.76 \\
    CAVP~\cite{chen2024unraveling}     & 62.31 & 78.46 & 13.56 \\
    \textbf{Ours} & \textbf{65.73} & \textbf{79.1} & \textbf{12.89} \\
    \midrule
    \multicolumn{4}{c}{\textbf{\textit{Dataset: VPO-MS}}} \\
    \midrule
    TPAVI~\cite{zhou2022audio}    & 54.3  & 71.95 & 22.45 \\
    AVSegformer~\cite{gao2024avsegformer} & 58.33 & 74.28 & 22.13 \\
    CAVP~\cite{chen2024unraveling}     & 64.31 & 78.92 & 18.67 \\
    \textbf{Ours} & \textbf{66.21} & \textbf{79.23} & \textbf{18.12} \\
    \midrule
    \multicolumn{4}{c}{\textbf{\textit{Dataset: VPO-MSMI}}} \\
    \midrule
    TPAVI~\cite{zhou2022audio}    & 51.73 & 68.85 & 27.65 \\
    AVSegformer~\cite{gao2024avsegformer} & 54.22 & 70.39 & 25.51 \\
    CAVP~\cite{chen2024unraveling}     & 60.36 & 75.6  & 22.12 \\
    \textbf{Ours} & \textbf{62.65} & \textbf{76.12} & \textbf{21.95} \\
    \bottomrule
  \end{tabular}
\end{table}
\begin{figure}[t]
  \centering
  \includegraphics[width=\linewidth, height=0.5\linewidth]{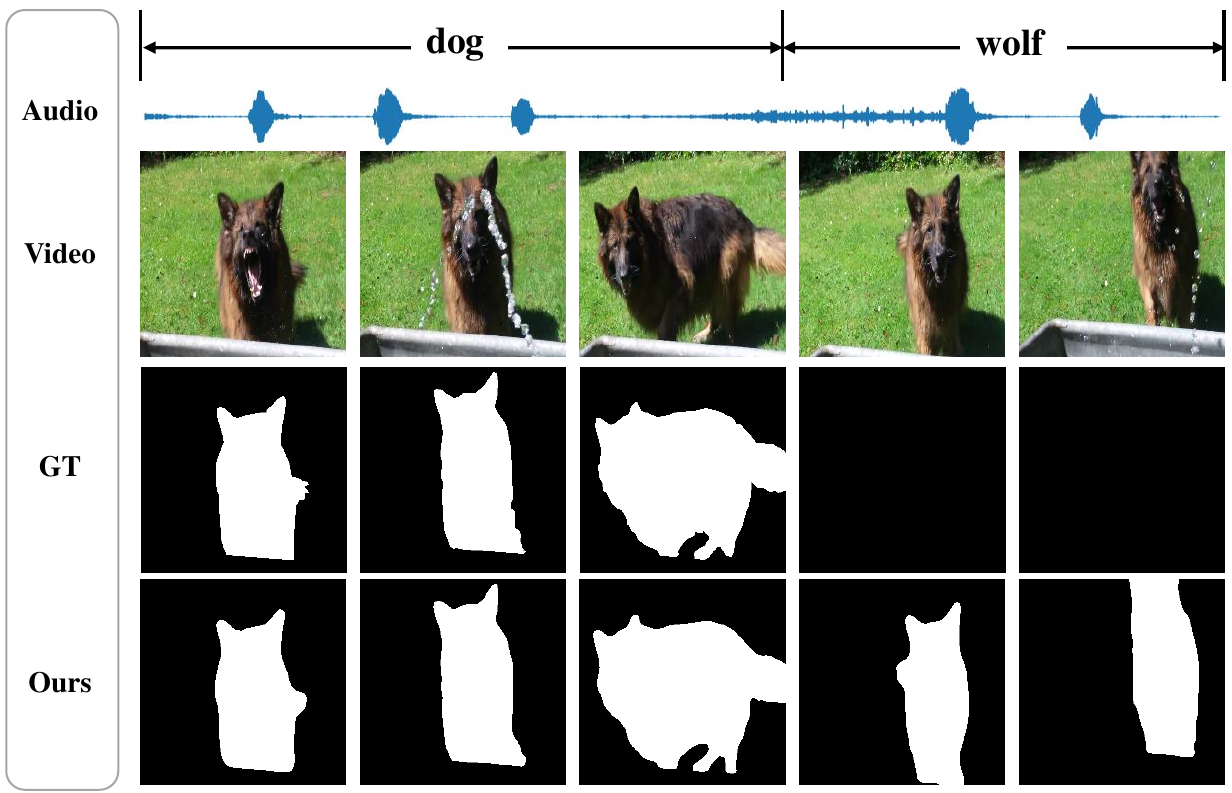}
  \caption{Failure cases: Our model demonstrates limitations when off-screen wolf howling coexists with silent dog, leading to incorrect segmentation.}
  \label{lim}
\end{figure}
\subsection{Limitation}
While our model achieves good performance in challenging multi-source scenarios, some failures as illustrated in Fig.~\ref{lim} should not be ignored as they can help reveal opportunities for improvement. Specifically, for visually similar and acoustically different objects, we may incorrectly segment objects that do not produce sound. For example, in certain video frames, the dog remains quiet, while the audio contains wolf howling. In such cases, our model tends to incorrectly associate the dog with the wolf howling, leading to incorrect segmentation. This issue has also been encountered in previous studies~\cite{zhou2022audio, li2023catr}, posing significant challenges to the application of AVS. We attribute the issue to the lack of relevant data in the AVS datasets, in which the sounding objects are visible and have no deceptive sound. However, sounding objects often do not appear in real-life scenes. The model should possess the ability to discriminate between situations where these objects are present and where they are absent, in order to accurately reflect the context of the scene.
\section{Conclusion}
In this paper, we strive to confront the challenge of effectively integrating audio and visual modalities in complex scenes. We propose the SNRP module that suppresses audio noise while enhancing relevant audio information. To model interactions between the audio and visual modalities, we further propose the DAMF model, allowing for greater adaptability to more complex scenarios. Extensive quantitative evaluations and qualitative visualizations demonstrate that our proposed method achieves state-of-the-art performance on benchmark AVS datasets. We hope these findings can inspire further exploration of noise-robust and perception-consistent modeling within multimodal research, especially in advancing audio-visual segmentation.


\bibliographystyle{IEEEtran}
\bibliography{sample-base}

\vspace{-1cm}
\begin{IEEEbiography}[{\includegraphics[width=25mm,height=35mm,clip,keepaspectratio]{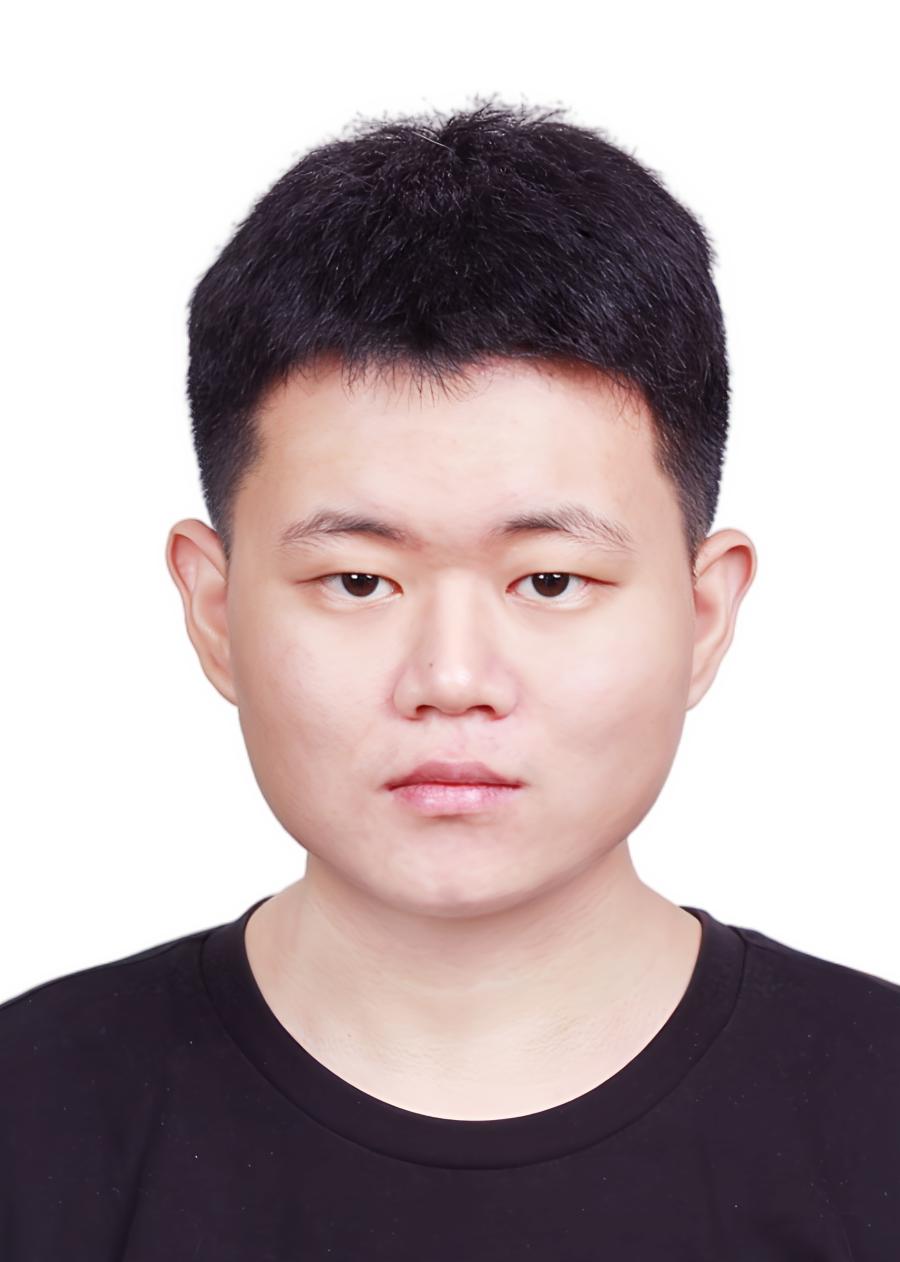}}]{Kai Peng}
	received the B.S. degree in computer science and technology from Xinjiang University in 2024. 
	
	He is currently with IIAU-OIP Lab, Dalian University of Technology, Dalian, China. His current research interests include computer vision, Audio-Visual segmentation and multimodal learning.
\end{IEEEbiography}
\vspace{-0.5cm}
\begin{IEEEbiography}[{\includegraphics[width=25mm,height=35mm,clip,keepaspectratio]{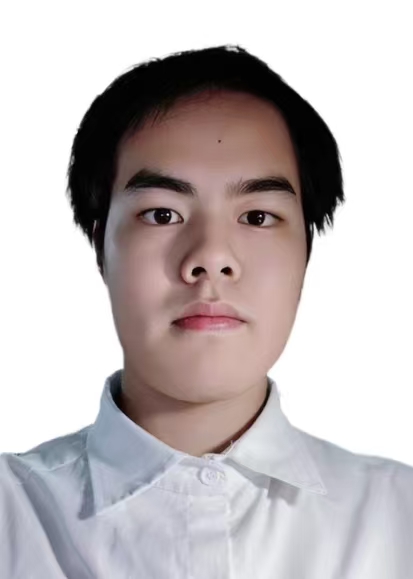}}]{Yunzhe Shen}
	received the B.S. degree in software engineering from Dalian University of Technology, Dalian, China, in 2024. 

    He is currently with IIAU-OIP Lab, Dalian University of Technology, Dalian, China. His current research interests include computer vision and multimodal learning.
\end{IEEEbiography}
\vspace{-0.5cm}
\begin{IEEEbiography}[{\includegraphics[width=25mm,height=35mm,clip,keepaspectratio]{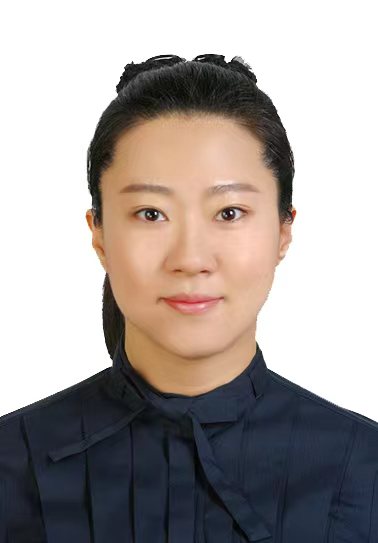}}]{Miao Zhang}
	(Member, IEEE) received the B.S. degree in computer science from the Memorial University of Newfoundland, St. John’s, NL, Canada, in 2005, and the Ph.D. degree in electronic engineering from Kwangwoon University, Seoul, South Korea, in 2012. From 2013 to 2015, she was an Assistant Professor with the Department of Game and Mobile Contents, Keimyung University, Daegu, South Korea, and an Adjunct Professor with the Department of Computer Science, DigiPen Institute of Technology, Redmond, WA, USA, respectively.
	
	She is currently an Associate Professor with the Key Laboratory for Ubiquitous Network and Service Software of Liaoning Province, DUT-RU International School of Information Science and Software Engineering, Dalian University of Technology, Dalian, China. Her research interests include computer vision, machine learning, and 3D imaging and visualization.
\end{IEEEbiography}
\vspace{1cm}
\begin{IEEEbiography}[{\includegraphics[width=25mm,height=35mm,clip,keepaspectratio]{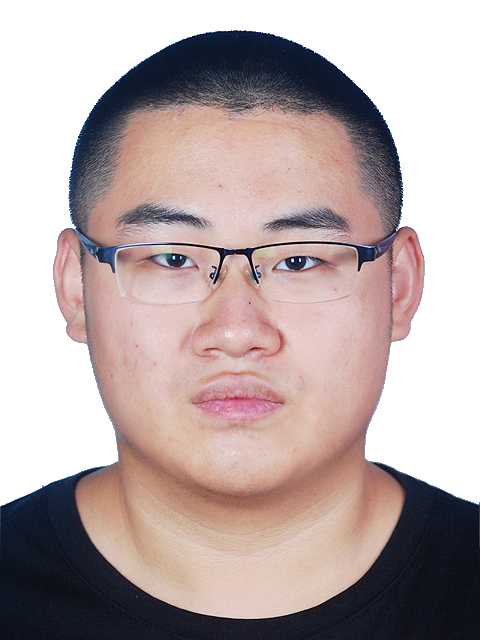}}]{Leiye Liu} 
    received the bachelor's degree from Dalian University of Technology. 
    
    He is currently a Ph.D. Student at Dalian University of Technology. His research interests include segmentation, state space model, foundation model, and multimodal perception.
\end{IEEEbiography}
\vspace{-0.5cm}
\begin{IEEEbiography}[{\includegraphics[width=25mm,height=35mm,clip,keepaspectratio]{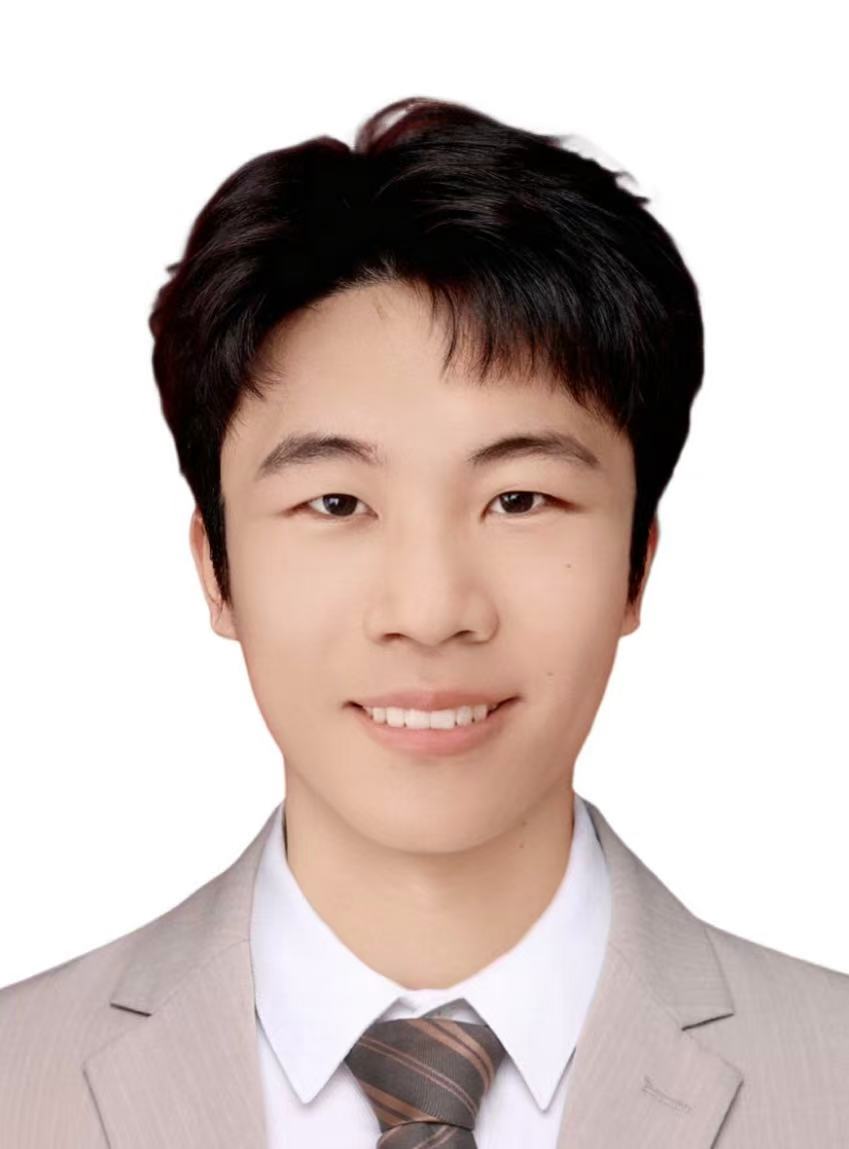}}]{Yidong Han}
	received the B.S. degree in  digital media technology from Dalian University of Technology in 2025. 
	
	He is currently with IIAU-OIP Lab, Dalian University of Technology, Dalian, China. His current research interests include computer vision, Audio-Visual segmentation and multimodal learning.
\end{IEEEbiography}
\vspace{-0.5cm}
\begin{IEEEbiography}[{\includegraphics[width=25mm,height=35mm,clip,keepaspectratio]{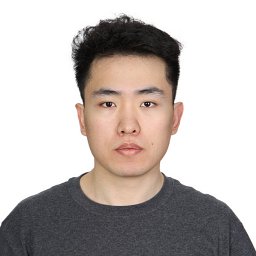}}]{Wei Ji}
    received the Ph.D. degree from the University of Alberta. He was a Visiting Ph.D. Student at Johns Hopkins University. He is currently a Postdoctoral Researcher with Yale University. 
    
    His research interests include segmentation, visual saliency, foundation model, multimodal perception, and medical image analysis. He achieved the CVPR Best Paper Candidate and the MICCAI Young Scientist Award Nominee.
\end{IEEEbiography}
\vspace{-0.5cm}
\begin{IEEEbiography}[{\includegraphics[width=25mm,height=35mm,clip,keepaspectratio]{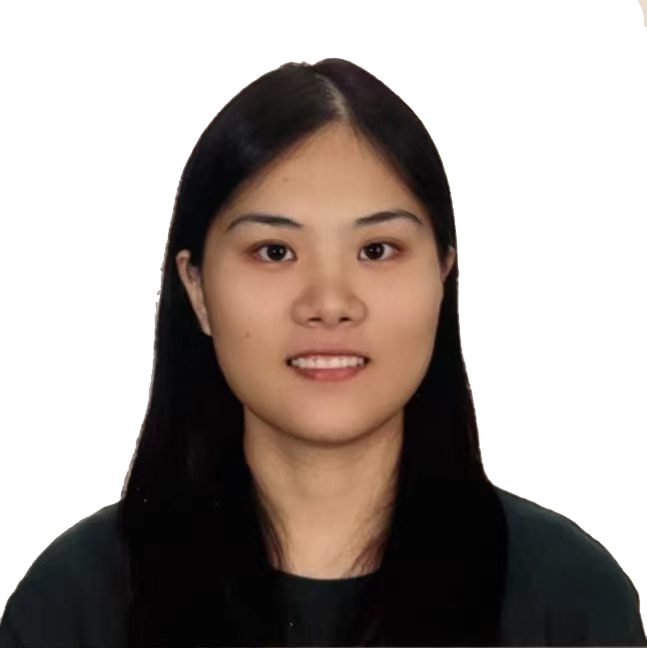}}]{Jingjing Li}
    is currently a Ph.D. degree candidate at University of Alberta, Canada.
    
    Her research interests include designing deep neural networks and applying deep learning in various fields of low-level vision, such as RGB salient object detection, RGB-D salient object detection, video object segmentation, and medical image segmentation.
\end{IEEEbiography}
\vspace{-0.5cm}
\begin{IEEEbiography}[{\includegraphics[width=25mm,height=35mm,clip,keepaspectratio]{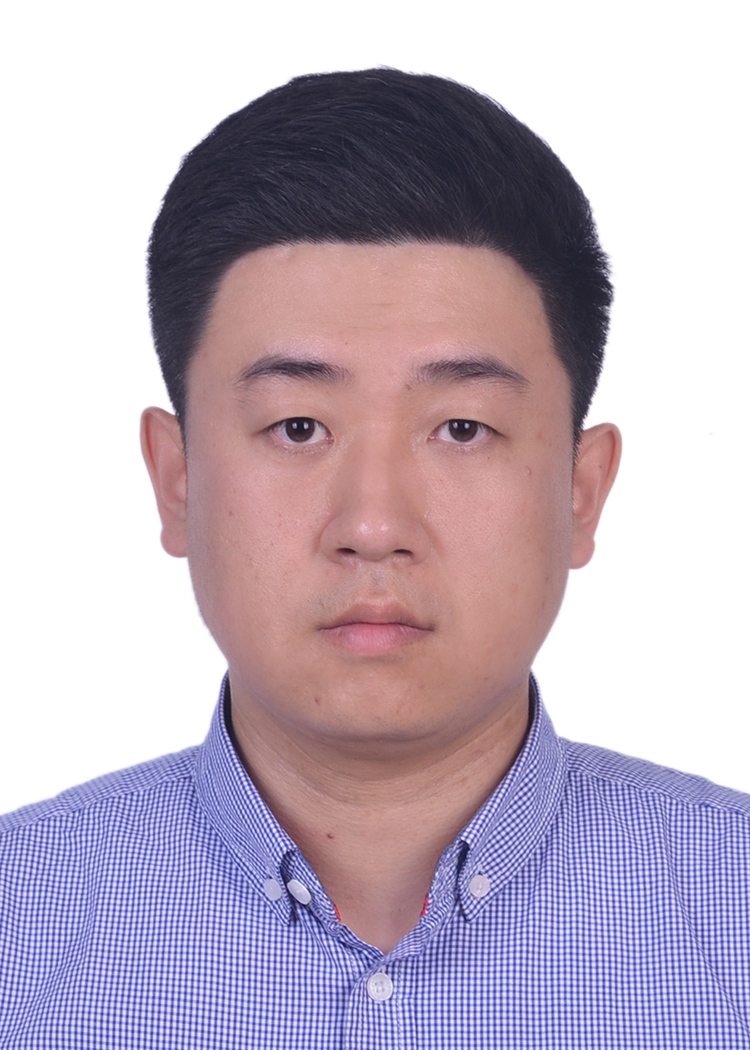}}]{Yongri Piao}
	(Member, IEEE) received the M.Sc. and Ph.D. degrees in information and communication engineering from Pukyong National University, Busan, South Korea, in 2005 and 2008, respectively. 
	
	Since 2012, he has been an Associate Professor of information and communication engineering with the Dalian University of Technology, Dalian, China. His research interests include 3D computer vision and sensing, object detection and target recognition, and 3D reconstruction and visualization.
\end{IEEEbiography}
\vspace{-0.5cm}
\begin{IEEEbiography}[{\includegraphics[width=25mm,height=35mm,clip,keepaspectratio]{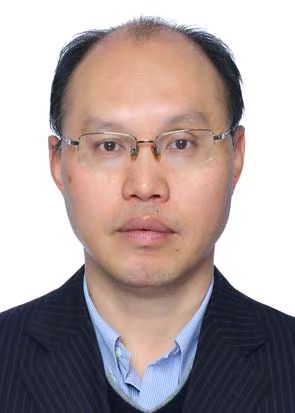}}]{Huchuan Lu}
	(Fellow, IEEE) received the M.S. degree in signal and information processing and the Ph.D. degree in system engineering from the Dalian University of Technology (DUT), Dalian, China, in 1998 and 2008, respectively. 
	
	He joined the Faculty of the School of Information and Communication Engineering, DUT, in 1998, where he is currently a Full Professor. His research interests include computer vision and pattern recognition with a focus on visual tracking, saliency detection, and segmentation.
\end{IEEEbiography}

\vfill

\end{document}